\documentclass[a4paper,fleqn, review]{cas-dc}

\usepackage[square,sort,comma,numbers]{natbib}
\setcitestyle{square}
  \usepackage{todonotes}
\DeclareUnicodeCharacter{2061}{}
\usepackage{tikz}
\usetikzlibrary{shapes.geometric, arrows}
\usepackage{subfig}
\usepackage{balance, todonotes, enumitem, graphicx}

\usepackage{amsmath}

\usepackage{algpseudocode}
\usepackage{algorithm}
\usepackage{bm}
\algdef{SE}[VARIABLES]{Variables}{EndVariables}
   {\algorithmicvariables}
   {\algorithmicend\ \algorithmicvariables}
\algnewcommand{\algorithmicvariables}{\textbf{Variables}} 

\usepackage{caption}
\def\tsc#1{\csdef{#1}{\textsc{\lowercase{#1}}\xspace}}
\tsc{WGM}
\tsc{QE}

\begin{document}
\let\WriteBookmarks\relax
\def\floatpagepagefraction{1}
\def\textpagefraction{.001}

\shorttitle{Bayesian Active Learning for Comparative Judgement: Estimating Reliability and Managing Multiple Criteria}


\title [mode = title]{Bayesian Active Learning for Comparative Judgement: Reliability Estimation and Multi-Criteria Evaluation with Applications in Educational Assessment 
}




%

\ifdefined\DOUBLEBLIND
    \shortauthors{Anonymous \textit{et al.}}    
    \author[]{-- Author Names Removed for Peer-Review --}
\else
    \shortauthors{Gray \textit{et al.}}

\author[1,2]{Andy Gray}[type=editor,
    orcid=0000-0002-1150-2052,
    twitter=codingWithAndy,
]

\cormark[1]


\ead{a.gray2@bathspa.ac.uk}
\ead{445348@swansea.ac.uk}



\affiliation[1]{organization={Bath Spa University},
            city={Bath},
            country={United Kingdom}}
\affiliation[2]{organization={Swansea University},
            city={Swansea},
            country={United Kingdom}}

\author[2]{Alma Rahat}[
    orcid=0000-0002-5023-1371,
    twitter=AlmaRahat,
    ]

\ead{a.a.m.rahat@swansea.ac.uk}

\author[2]{Tom Crick}[
    orcid=0000-0001-5196-9389,
    twitter=ProfTomCrick,]


\ead{thomas.crick@swansea.ac.uk}




\author[3]{Stephen Lindsay}[orcid=0000-0001-6063-3676]
\affiliation[3]{organization={University of Glasgow}, 
            city={Glasgow},
            country={United Kingdom}}

\ead{stephen.lindsay@glasgow.ac.uk}

\fi

\begin{abstract}
    Comparative Judgement (CJ) offers an alternative approach to assessment by focusing on the holistic evaluation of a piece of work, rather than dissecting it into discrete components that contribute to the overall decision-making. This enables evaluators to consider the overall quality and coherence of a submission, leveraging the human ability to make nuanced comparisons and supporting more reliable and valid assessments. By emphasising the overall impression, CJ aligns more closely with real-world evaluations, where the interplay of various elements determines impact and effectiveness.
    
    Bayesian Comparative Judgement (BCJ) with active learning — a relatively recent innovation — allows preferences to be modelled directly and uncertainty to be systematically reduced by selecting the most informative pairs for evaluation. However, BCJ lacks a reliability measure akin to the well-established Scale Separation Reliability (SSR) used in Bradley-Terry-based CJ models. Furthermore, while multiple criteria often underpin holistic decisions in CJ (e.g. rubric-based assessments in education), these are rarely made explicit, even when the weights for each criterion are known. This restricts the ability to assess decisions from distinct perspectives or offer focused feedback without individually revisiting and analysing each item — a process that greatly increases cognitive load and diminishes one of CJ’s core strengths.
    
    In this paper, we address these limitations through a Bayesian extension of CJ. We introduce two new reliability metrics — Mode Agreement Percentage (MAP) and Expected Agreement Percentage (EAP) — derived from preference distributions in BCJ. These metrics quantify assessor agreement at the pairwise level and help identify contentious comparisons. We further extend BCJ to handle multiple \textit{independent} learning outcome (LO) components, as defined in rubrics, enabling both component-wise and holistic predictive rankings with associated uncertainty estimates via mixture distributions. Additionally, we propose a multi-criteria entropy measure to guide the selection of the most informative comparisons. \textbf{Through experiments with real-world datasets from educational assessments, we demonstrate the effectiveness of the proposed methods.}
\end{abstract}





\maketitle

\section{Introduction}
    \label{sec:intro}

    Humans are naturally better at making relative estimations than assigning absolute values \cite{thurstone1927law, laming1984relativity}. This can be best understood through an intuitive example: imagine someone is asked to estimate the weight of a cow. Without significant experience in weighing cows, they might struggle to provide an accurate figure. However, if asked to compare two cows and identify which one is heavier, they are likely to perform the task more confidently and with greater accuracy.
    
    In 1927, Thurstone \cite{thurstone1927law} was the first to introduce a formal model for ranking a collection of items — also referred to as representations or stimuli — based solely on pairwise comparisons. In this \textit{comparative judgment} (CJ) framework, a judge (or assessor) simply selects the item of higher quality from each pair. Since then, extensive research has been devoted to the problem of inferring rankings from such comparisons. Among the various models developed, the Bradley–Terry model (BTM) \cite{bradley1952rank} remains the most widely adopted in practice.
    
    For uncertainty quantification, the conventional approach is frequentist, where scores are estimated by maximising the likelihood function \cite{hunter2004mm}. Nonetheless, Bayesian alternatives \cite{caron2012efficient} have been gaining traction. Regardless of the methodological stance, most models aim to infer latent scores that underpin the rankings, often relying on distributional assumptions (e.g. Normality) -- assumptions that continue to be questioned \cite{ballinger1997decisions,kelly2022critiquing}.
    
    Crucially, these scores are never directly observed; only the pairwise comparisons are available. This raises concerns about the interpretability and justification of the inferred scores. In essence, the primary outcome of interest is often the ranking itself, rather than the underlying scores.
    
    The core premise in CJ is that we typically rely on human input to identify the `winner' in a pairwise comparison. Naturally, this constrains the number of comparisons that can be feasibly collected. To address this limitation, active learning strategies, where the outcomes of previous comparisons guide the selection of subsequent pairs in order to maximise information gain about the overall ranking, should be deployed. However, in practice, a principled mechanism for pair selection is often absent, leading researchers to rely on random sampling, `round-robin', or heuristic-based methods \cite{pollitt2012method, bramley2015investigating}.
    
    Although active learning is highly relevant in this context, it was first formally addressed by Jamieson et al. \cite{jamieson2011active} in 2011, under the assumption of deterministic comparisons — meaning that repeated evaluations of the same pair consistently yield the same winner. Later studies, such as those by Heckel et al. \cite{heckel2019active}, introduced stochastic models to better reflect the variability in human binary decision-making \cite{tversky1969substitutability}. Nonetheless, these approaches have largely remained within the frequentist framework for estimating rankings.
    
    A recent development in CJ is the introduction of Bayesian Comparative Judgement (BCJ) by Gray et al. \citep{GRAY2024100245}. In contrast to earlier models such as the BTM, which impose a likelihood on latent scores and estimate ranks via maximum likelihood, BCJ models pairwise preferences directly as outcomes of Bernoulli trials. This formulation avoids several assumptions that are often considered unrealistic in practical assessment settings.
    
    A key strength of BCJ is its ability to naturally handle imperfect and stochastic judgements — whether from individual or multiple assessors — by encoding uncertainty through a Beta distribution for each pair. These distributions allow for a probabilistic estimation of an item’s rank distribution, conditioned on the observed comparisons.
    
    In addition, the authors introduce an entropy-based active learning strategy to identify the next most informative pair to present to assessors -- a novel contribution within the Bayesian framework. Empirical evaluations show that BCJ consistently outperforms existing BTM-based methods across both synthetic and real-world datasets.
    
    Beyond validity -- that is, the empirical accuracy of the ranking process in CJ -- an equally important consideration is reliability, typically defined as the consistency of judgements for the same pair of items \cite{ashton2000review}. In binary decision-making, it is well established that outcomes are stochastic \cite{tversky1969substitutability}, meaning that a judge may not always be consistent with their own choices, and different judges may also disagree. Such intra- and inter-assessor variability is natural and, in extreme cases, may even violate transitivity properties such as the triangle inequality (e.g. if a is preferred to b, and b to c, then a should be preferred to c) \cite{ballinger1997decisions}. It is therefore essential not only to demonstrate general agreement across assessors for each pairwise decision but also to quantify the extent of such agreement.
    
    A commonly used metric for reliability in CJ is scale separation reliability (SSR), which can be interpreted as a measure of how well the observed scores (derived from fitting the BTM) align with the expected scores. SSR is typically computed by subtracting the standard error -- obtained from the maximum likelihood estimation -- from the observed scores \cite{kinnear2025comparative}. Although SSR can be difficult to interpret and has been shown to yield biased estimates \cite{bramley2015investigating}, it remains a useful indicator of inter-rater reliability.
    
    In the context of Bayesian Comparative Judgement (BCJ), validity is supported by the probabilistic framework, which can be empirically tested when target ranks are available. However, a robust and widely accepted measure of agreement for pairwise decisions within the Bayesian setting has yet to be established.
    
    It is also important to recognise that while pairwise comparisons in comparative judgement (CJ) are typically made holistically, researchers generally acknowledge the presence of multiple latent attributes that influence decision-making \cite{yu2022multidimensional}. As early as 1959, Hefner extended the Thurstone model to explore this idea \cite{hefner1959extensions}, and subsequent research has focused on estimating the contributions of underlying latent factors from pairwise comparisons alone.
    
    In many real-world judgement scenarios, however, the relevant criteria and their contributions are explicitly defined in advance. For instance, in educational assessment, students are evaluated using a predefined rubric that specifies distinct criteria, each contributing to the final mark with predetermined weights. In such cases, it would be natural to collect pairwise comparisons at the level of individual criteria. Yet, to the best of our knowledge, this approach has not been explored within the CJ framework, which has traditionally emphasised holistic comparisons.
    
    Although some existing methods attempt to infer latent dimensions from holistic choices, it may be difficult to align these inferred dimensions with specific, predefined criteria. Moreover, they lack a principled mechanism for aggregating criterion-level decisions into a single, coherent ranking.
    
    To address these limitations, this paper builds on BCJ and makes the following key contributions:
    \begin{itemize}
        \item We introduce two novel metrics -- Mode Agreement Percentage (MAP) and Expected Agreement Percentage (EAP) -- based on a Beta prior over pairwise preferences. These metrics quantify the level of agreement among assessors and help identify controversial comparisons. In particular, EAP serves as a direct indicator of reliability under uncertainty due to limited data and offers a principled stopping criterion for data collection.
        \item We propose new methods for estimating overall ranks and associated predictive uncertainties from pairwise comparisons made per criterion. This framework, which we term Multi-Criteria BCJ (MBCJ), enables criterion-specific ranking and uncertainty estimation.
        \item We show how a holistic entropy can be calculated to drive the selection of the most informative pair to be evaluated next in MBCJ.
        \item We demonstrate, for the first time, that MBCJ performs comparably to standard BCJ in experiments using real assessment data, while providing finer-grained insights into item preferences across individual criteria.
    \end{itemize}
    
    We structure the remainder of the paper as follows. In Section \ref{sec:related_work}, we introduce the educational assessment context, review related work, and summarise the key concepts behind Bayesian Comparative Judgement (BCJ). Section \ref{sec:reliability} outlines our proposed reliability measure. In Section \ref{sec:mbcj}, we extend BCJ and the associated active learning approach to incorporate multiple criteria. Section \ref{sec:model,data,metrics} details our experimental setup. We present and discuss our results in Section \ref{sec:experiment_discussion}, and conclude with a summary and future research directions in Section \ref{sec:conclusion}.

\section{Related Work and Background}
    \label{sec:related_work}
        
        \subsection{Rubric-Based Educational Assessment}

        Learning is a fundamental aspect of life, and teaching remains one of society’s most vital responsibilities. While the teaching and learning process can be demanding, it is also deeply fulfilling. A key component of effective teaching is the ability to assess and monitor student performance. This is essential for purposes such as reporting, setting ability levels, identifying students who may need additional support, and offering feedback to help them improve \citep{wellington2007secondary, yeomans2013teaching}. As such, marking is an integral part of the teaching process \citep{brooks2019preparing}. One of the most widely adopted tools for this purpose is the marking rubric, which outlines multiple assessment criteria or dimensions in a clear and structured manner.

        \begin{figure*}[b]
            \centering
            \includegraphics[width=\textwidth]{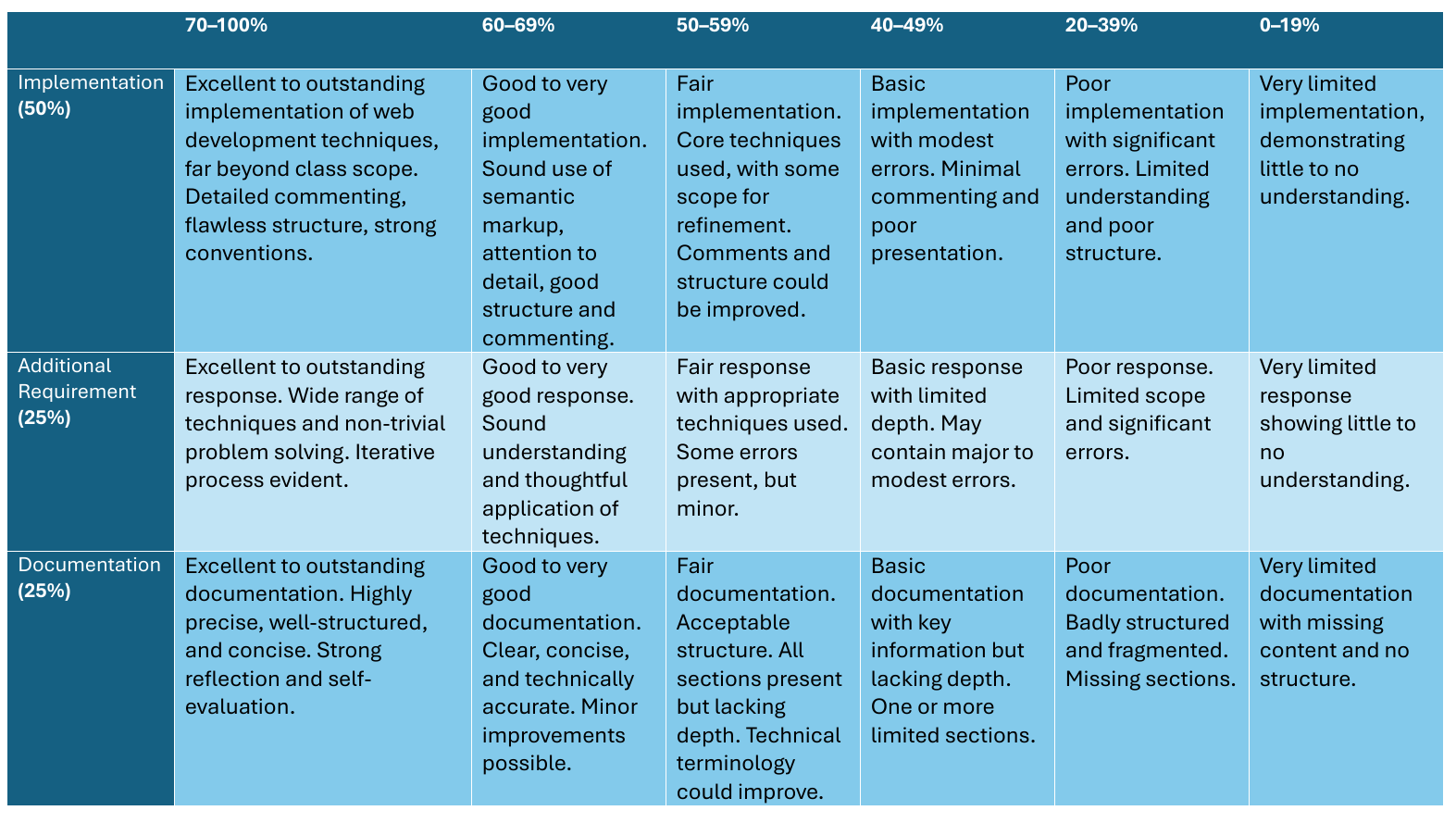} 
            \caption{An example marking rubric for a level 4 undergraduate module offered at the Bath Spa University, UK. It provides an overview of the quality required to achieve a certain grade (along the columns), based on different criteria (along the rows) for the assessment as designed by the assignment owner. Here, the criteria are Implementation, Additional requirements, and Documentation.}
            \label{fig:msc-rubric}
        \end{figure*}

        Among the various strategies educators use to assess student achievement, rubric marking has gained prominence as a transformative approach that goes beyond traditional grading systems \citep{Cox2015The}. It has become a cornerstone of modern assessment, offering a transparent and consistent framework that benefits both teachers and learners. Rubrics help clarify expectations and assessment standards, enabling students to better understand what is required to succeed in their academic tasks \citep{Poh2015A, Cox2015The}. Their built-in consistency promotes faster, fairer, and more reliable grading—an essential factor in upholding the integrity of educational evaluation \citep{Ragupathi2020Beyond}.
        
        Rubrics, grounded in well-defined criteria -- often aligned with learning outcomes (LOs) -- and explicit quality benchmarks (with associated weight of contributions to the overall marks), provide a systematic method for assessing students’ knowledge and their ability to apply it meaningfully \citep{olson2021rubrics}. In today’s education system, which values holistic growth and deep learning, rubrics have become indispensable tools for educators \citep{cox2015rubric}. They support more comprehensive and insightful evaluations, while also enabling constructive feedback, tailored learning pathways, and the development of well -- rounded individuals ready to navigate the complexities of the 21st century \citep{sambell2019assessment}.
        
        A marking rubric -- also known as a scoring rubric -- is a structured tool that clearly communicates assignment expectations by listing assessment criteria and describing different levels of performance; see, for example, Figure~\ref{fig:msc-rubric}. It offers a transparent and objective approach to evaluating student work across a range of formats, including essays, group projects, creative outputs, and oral presentations. Rubrics can be applied to any task where students are expected to demonstrate their learning.

        Rubrics have several advantages, such as providing clarity and consistency in grading \citep{Jonsson2007The}. They offer clear expectations and grading criteria to students, which can assist them in understanding what is required to excel in an assignment \citep{Cockett2018The}. They can make grading much quicker, more consistent, and fair \citep{Jonsson2007The}. Furthermore, rubrics can provide students with informative feedback on their strengths and weaknesses so that they can reflect on their performance and work on areas that need improvement \citep{Reddy2010A}. Rubrics also encourage learners to develop critical thinking about their own scores and work \citep{Reddy2010A}. However, rubrics also have their drawbacks. The language of rubrics is not always as clear as it is supposed to be, which adds to their complexity \cite{Panadero2020A}. The lower scale may use negative terms to describe student performance, which may discourage the learners. Some opponents of rubrics feel they are more subjective than a letter grade.
        
        In higher education, rubrics have been recognised for enhancing student self-assessment, self-regulation, and understanding of assessment criteria \citep{Cockett2018The}. However, some students perceive rubrics as restrictive and associate them with increased stress related to assessments \citep{Cockett2018The}. The involvement of students in the design and implementation of rubrics is essential for their success \citep{Cockett2018The}. In primary education, particularly in the teaching and assessment of mathematical reasoning, rubrics have been found to improve teachers’ diagnostic skills and indirectly influence their use of formative feedback \citep{Smit2017Effects}. However, the direct effects on student self-assessment are more apparent than the effects on student outcomes, highlighting the need for further research into the mediated effects of self-regulation and self-efficacy \citep{Smit2017Effects}.
                
        Empirical data from higher education indicates increased use, driven by the demands for consistency and transparency in assessment \citep{Hack2015Analytical}. While the reliability of rubrics is supported by evidence, the impact on student learning necessitates further robust evaluation \citep{Hack2015Analytical}. Ultimately, rubrics are invaluable tools in educational assessment, with their effectiveness contingent upon their design, implementation, and the context in which they are used. The potential of rubrics is vast, yet challenges remain that require ongoing research to understand and address fully. When effectively implemented, rubric marking can significantly enhance the reliability and validity of assessments, positively influencing student learning and performance. However, the actual impact of rubric marking varies depending on specific contexts and implementations, and it is influenced by factors such as the clarity of criteria, assessor training, and the feedback provided to students. These general pros and cons underscore the need for a nuanced application of rubrics in educational settings.

    \subsection{Comparative Judgement}
        \label{subsec:cj}
        CJ is a technique used to derive ranks from pair-wise comparisons. The concept of CJ is used in academic settings to allow teachers to compare two pieces of work and select which is better 
        in a holistic manner. After each comparison, another pair is selected. This is repeated until enough pairs have been compared to generate a ranking of the work marked. We detail a typical CJ process in Algorithm~\ref{alg:opt} \cite{GRAY2024100245}.

        \begin{algorithm} [h!]
          \caption{Standard comparative judgement procedure.}
          \label{alg:opt}
          \textbf{Inputs.}
          \begin{algorithmic}[]
            \State $N:$ Number of items.
            \State $K:$ Multiplier for computing the budget for the number of pairs to be assessed.
            \State $I:$ Set of items.
          \end{algorithmic}
          \bigskip
          \textbf{Steps.}
          \begin{algorithmic}[1]
            \State $B \gets N \times K$ \Comment{\small{Compute the budget.}}
         \State $G \gets \langle \rangle$ \Comment{\small{Initialise list of selected pairs.}}
         \State $W \gets \langle \rangle$ \Comment{\small{Initialise list of  winners.}}
         \State $\mathbf{r} \gets \left(\frac{N}{2}, \dots, \frac{N}{2}\right)^\top ~|~ 
         \lvert \mathbf{r} \rvert = N$  \par
                \hskip\algorithmicindent\Comment{\small{Initialise rank vector with mean rank for all items.}}
        	\For{$b = 1 \rightarrow B$}
                    \State $(i, j) \gets \text{SelectPair}(I)$ \label{alg:sel_pair}
                    \Comment{\small{Pick a pair of items.}} \label{alg:pick_pair}
                    \State $G \gets G \oplus \langle (i,j) \rangle$ \Comment{\small{Append the latest pair.}}
                    \State$ w \gets \text{DetermineWinner}(i, j) $\Comment{\small{Pick a pair of items.}}
                    \State $W \gets W \oplus \langle w \rangle$ \Comment{\small{Append the latest winner.}}
                    \State $\mathbf{r} \gets \text{GenerateRank}(G, W) $ \Comment{\small{Update rank vector.}} \label{alg:gen_rank}
        	\EndFor
        	\State \Return $\mathbf{r}$
          \end{algorithmic}
        \end{algorithm}
    
        An important benefit to CJ within an academic setting is reducing the teacher's cognitive load~\citep{chen-et-al:2023}, as comparing two pieces of work is faster than marking each individual piece of work, while also insisting the teacher is being non-biased towards a student and consistent~\citep{sadler:1989}. This is difficult to achieve \citep{bramleypaired:2007}, and CJ helps, to an extent, address this challenge; for further discussion of this, we refer to the following literature where 
        the teachers can be referred to as the judges~\citep{,benton2018comparative, bartholomew2019using, christodoulou2017making}.

        CJ is based on Thurstone's proposed technique in 1927, known as `the law of comparative judgement' \citep{thurstone1927law}. Thurstone discovered that humans are better at comparing things to each other rather than making judgements in isolation. 
        Therefore, he proposed making many pair-wise comparisons until a rank order has been created~\citep{thurstone1927law, benton2018comparative, bartholomew2019using}. Pollitt \textit{et al.} played a crucial role in introducing and popularising 
        it within an education setting \citep{pollitt1996raters, pollitt2004let}.

        A growing body of evidence supports using CJ as a reliable alternative for assessing open-ended and subjective tasks. The judgements recorded by teachers, more generally termed \textit{raters} or \textit{judges}, are fed into a BTM (see 
        \citep{GRAY2024100245, gray2022using}
        for more details on the BTM) to produce scores that represent the underlying quality of the scripts \citep{bradley1952rank, luce1959individual}. These scores have the appealing property of being equivalent across comparisons \citep{andrich1978rating}.  

        A key justification for using CJ within the educational assessment process is that the rank orders it produces tend to have high levels of reliability. For example, in 16 CJ exercises conducted between 1998 and 2015, the SSR indices, which is equivalent to Cronbach’s alpha, a measure of internal consistency and scale reliability \citep{verhavert2018scale}. The correlation coefficient scores were between $0.73$ to $0.99$ compared to rubric-based grades \citep{steedle2016evaluating}. With a correlation coefficient of $1.0$ representing perfect agreement, an SSR score of $0.70$ or above is typically considered high enough to proclaim strong agreement~\citep{hinkle2003applied}.

        To the best of our knowledge, in a multi-criteria aspect, CJ’s potential in this area has been researched once, with two criteria where pairwise comparisons were used to rank exemplar scripts required for later script evaluation \citep{mcgrane2018applying}. McGrane et al.’s study aimed to expand the traditional use of CJ to incorporate a two-staged process, using CJ to generate calibrated exemplars followed by matching exemplars to performances. The study evaluated performances across two tasks — narrative and persuasive writing — and comprised performances from two calendar years of administration. Judgements were made using two different dimensions, which they referred to as writing conventions and authorial choices criteria. However, the rankings for the different dimensions were independent and not combined to create an overall score and rank for the items being compared. It was used to create a sample scale as a source of 36 calibrated exemplars for the second part of their experiment, where they then used these exemplars to match the remaining items to the most similar item in the calibrated exemplars. So, there is a clear gap in the literature in the use of CJ for multi-criteria pairwise comparisons where the criteria are known, and weighted aggregation of ranks to produce overall ranks while driving the selection of pairs in an informed manner.

        The statistical methods for CJ mainly vary in two steps of Algorithm \ref{alg:opt}: selecting the next pair to evaluate in line \ref{alg:sel_pair} and generation of rank in line\ref{alg:gen_rank}. Below, we first discuss the standard approach with BTM as a central model for CJ, and then shed light on typical stopping criterion in this context.
    
        \subsubsection{Pair Selection Methods}
        
        A key consideration when implementing a CJ approach to marking is how to select the next pair of items for evaluation (see line~\ref{alg:pick_pair} in Algorithm~\ref{alg:opt}) and gather data on which item is judged superior. Various methods exist for generating pair suggestions; see, for example,~\cite{jones_davies_2022}. These are often developed on an ad hoc basis. Broadly, two commonly used strategies are random selection and the No Repeating Pairs (NRP) method, also known as round-robin.

        The random approach selects each pair uniformly at random from the set of all possible combinations until the evaluation budget is exhausted. Although this can result in the same pair being presented more than once, such repetition is unlikely when the number of items $N$ is large. This method resembles a known random search strategy and is particularly effective in large-scale scenarios~\cite{bergstra2012random}. It remains the most widely adopted technique~\cite{jones_davies_2022, benton2018comparative}.
        
        Alternatively, the NRP method ensures that no pair is repeated until all possible combinations have been evaluated~\cite{jones_davies_2022, ofqual2017}. This guarantees that each item is seen an equal number of times, although the specific pairings are still selected randomly. While this prevents early repetition, it does not account for uncertainty in the comparisons, meaning that some pairs may be evaluated even when the difference between them is already clear.

        \subsubsection{Rank Generation}

            In this section, we briefly discuss the rank generation approach in the popular BTM method. Given a set of $N$ indices, $I = \{1, \dots, N\}$, where each element $i$ is an index to the $i$th item of the $N$ items that we wish to rank. Typically, before ranking, we assume that there are latent scores for each item, $\bm{\gamma} = \{\gamma_1, \dots, \gamma_N\}$, where $\gamma_i \in [0,1]$ is a score representing the quality of the $i$th item; the higher, the better. With this, we can compute the probability that the $i$th item would beat the $j$th item in a comparison as follows: 
            \begin{align} 
            P(i \succ j) = \frac{\gamma_i}{\gamma_i + \gamma_j}. 
            \end{align} 
            
            Assuming independence of pairings between items, the log-likelihood for the performance vector $\bm{\gamma}$ is then defined as: 
            \begin{align} 
            \label{eq:btm_liklihood}
            L(\bm{\gamma}) = \sum_{i=1}^N \sum_{j=1,, i\neq j}^N \left[\omega_{[i,j]} \ln(\gamma_i) - \omega_{[i,j]} \ln(\gamma_i + \gamma_j)\right], 
            \end{align} 
            where $\omega_{[i,j]}$ is the number of times the $i$th item won a pairwise comparison against the $j$th item. An iterative maximisation-minorisation algorithm is then used to locate the best parameters for this parametric model \cite{hunter2004mm}. It is then straightforward to compute the expected rank for the items: $\bm{r}_{i \in I} = (N+1) - \arg\text{sort}(\bm{\gamma})$, where 1 is the top rank.
            
            \subsubsection{Stopping Criterion} 
            
            The traditional stopping criterion in CJ is typically defined as a budget on the number of pairwise evaluations. In this context, we assume a budget of $N \times K$, where $K$ is a multiplier commonly set to 10~\cite{jones_davies_2022}.
    
            Ofqual has noted that exceeding the optimal number of comparisons can reduce the effectiveness of the final ranking. However, the precise optimal number remains unknown~\cite{ofqual2017}. This limitation is more a consequence of the rank generation model than the pair selection strategy. This behaviour was demonstrated experimentally by Gray \textit{et al.}~\cite{GRAY2024100245}, and it highlights a key limitation of standard CJ. It should be noted that their Bayesian approach did not suffer with an increase in data; in fact, its estimates improved with additional comparisons. In the next section, we discuss Bayesian approaches to CJ.

    \subsection{Bayesian Comparative Judgement}
    \label{sec:bcj}

       The Bayesian approach is especially appealing due to its clear specification of prior beliefs about item rankings and its ability to propagate uncertainty through to the posterior. As more data becomes available, the overall uncertainty in our estimates decreases. This framework thus supports decision-makers in making more informed choices.
    
        In the literature, Bayesian versions of CJ typically rely on the BTM framework~\citep{wainer2022bayesian}. The estimation process usually begins with assumptions about prior distributions over $\gamma_i$ and $\omega_{[i,j]}$, corresponding to the likelihood function described in Equation \eqref{eq:btm_liklihood}. A Markov Chain Monte Carlo (MCMC) method is then employed to estimate the posterior distributions of these parameters and to compute the expected ranks of all items. However, this process can be computationally intensive, even for a moderate number of items, with complexity increasing alongside the number of items and comparisons. For instance, in an experiment using PyMC \cite{salvatier2016probabilistic} based on the work of Wainer et al. \cite{wainer2022bayesian}, ranking 10 items with 50 comparisons took approximately 10 seconds, whereas ranking 50 items with 250 comparisons required nearly eight times longer. Crucially, no active learning strategies have been developed for this approach, and therefore, we do not consider it further in this paper.
        
        The Bayesian approach — or BCJ \cite{GRAY2024100245} — explored in this paper is conceptually distinct: it avoids assumptions about unobservable latent scores and instead models directly the observed pairwise preferences. This makes the prior assumptions more transparent and less contentious. While estimating scores can offer advantages — such as quantifying the dispersion between items in addition to ranking them — analyses typically prioritise generating ranks over examining score dispersion. Acknowledging this, Gray \textit{et al.} derived item ranks directly from preference densities, eliminating the need to estimate latent scores. This results in a computationally efficient framework, where time complexity scales with the number of items but not with the number of comparisons. Moreover, the method supports an entropy-based strategy for actively selecting the most informative pairwise comparison to reduce overall uncertainty, which is necessary for an interactive system.
    
        It is worth noting that BCJ is well-suited for interactive systems. Even with a large number of items — for example, 300 — the method can identify the next pair to evaluate in approximately 10 milliseconds and generate the full rank distribution in about 15 seconds.
    
        In the following sections, we describe the rank generation and pair selection processes in BCJ, followed by a discussion of its limitations and how this paper addresses them.

    \subsubsection{Rank Generation}

        Following the proposed method in \cite{GRAY2024100245}, pairwise decision can be deemed as outcomes of a Bernoulli process, and as such, the likelihood over the proportion of wins when a pairwise comparison is performed between item $i$ and $j$ can be defined as follows: 
        \begin{align} 
            L(p | \bm{x}) = p^w \times (1-p)^{n-w}, 
        \end{align} 
        where the vector of decisions $\bm{x} = (x_1, \dots, x_n)^\top$ contains $n$ comparisons indicating whether $i$ was preferred over $j$, with $x_k = 1$ representing a win for $i$ and $x_k = 0$ indicates the opposite. Here, $w$ is the number of wins for $i$.
    
        In this framework for pairwise decisions, the conjugate prior for this likelihood is known to be a Beta distribution that has two shape parameters: $\alpha$ and $\beta$. With $\alpha_{\text{prior}} = 1$ and $\beta_{\text{prior}} = 1$ representing a prior of equally likely wins for item $i$ and $j$, the posterior after $n$ comparisons is derived by: 
        \begin{align} 
        \alpha_{\text{post}} &\gets \alpha_{\text{prior}} + w,\\ 
        \beta_{\text{post}} &\gets \beta_{\text{prior}} + (n-w). 
        \end{align}
    
        The probability density of preferring $i$ over $j$ is then defined as: 
        \begin{align}
            \pi(i \succ j) = \mathcal{B}(\alpha_{\text{post}}, \beta_{\text{post}}),        
        \end{align}
        where $\mathcal{B}(\cdot, \cdot)$ is the Beta distribution. With this, the probability that $i$ would win over $j$ is given by: 
        \begin{align}
        \label{eq:prob_pref}
            P(i\succ j) &= P(\pi(i \succ j) > 0.5) \nonumber\\
            &= 1 - \mathcal{F}(0.5~|~i \succ j),
        \end{align}
        where $\mathcal{F}(0.5~|~i \succ j)$ is the cumulative density function (CDF) for the Beta distribution.
    
        Gray \textit{et al.} derived a discrete probability distribution for the rank of item $i$, denoted as $P(r_i = a)$, where $a$ is an integer in the range $[1, N] \subset \mathbb{N}$. The specific formulation is omitted here, as it does not directly inform the discussion in the remainder of this paper. Readers interested in the full derivation are encouraged to consult \cite{GRAY2024100245}.
    
        With this, we can compute the expected rank for any item as $\mathbb{E}[r_i] = \sum_a aP(r_i = a)$. The ranks of the items can then be calculated as: $\bm{r}_{i \in I} = (N+1) - \arg\text{sort}(\mathbb{E}[\bm{r}])$.
    
    \subsubsection{Pair Selection}
    \label{bcj:pair-del}
        Gray \textit{et al.} adopted a straightforward uncertainty sampling approach, i.e., the next evaluations should be performed on the pair with the highest uncertainty, as it is expected to be the most informative for enriching the dataset~\cite{lewis1995sequential}. To quantify uncertainty, they used the \textit{entropy} of the Beta distributions that represent the current preferences between items. A higher entropy indicates greater uncertainty in the estimation. The entropy of a Beta distribution $\mathcal{B}(\alpha_{post}, \beta_{post})$ with shape parameters $\alpha_{post}$ and $\beta_{post}$ is given by~\cite{lazo1978entropy}:
        \begin{align}
         H\left[\pi(i\succ j)\right] = \ln \mathcal{B}(\alpha_{post},\beta_{post}) - (\alpha_{post} - 1) \psi(\alpha_{post}) \nonumber
         \\- (\beta_{post} - 1) \psi(\beta_{post}) + (\alpha_{post}+\beta_{post}-2)\psi(\alpha_{post}+\beta_{post}),
        \end{align}
        where, $\psi(\cdot)$ is the Digamma function. This approach was shown to be superior to the typical random and NRP pair selection methods.
    
    \subsubsection{Limitations}
    
        Although the BCJ method is attractive for its speed and accuracy, a key open question is how to assess its reliability. In this paper, we address this by examining inter- and intra-rater agreement as indicators of reliability. This also opens the door to defining a stopping rule for data collection — for instance, halting further evaluations once reliability exceeds a predefined threshold.
        
        Furthermore, many real-world tasks, such as rubric-based assessments, involve multiple evaluation criteria with predefined weights. In such cases, assessing items independently on each criterion may be more appropriate than relying on holistic comparisons. This approach can offer clearer insights into how items differ in quality. However, existing literature lacks methods for integrating decisions across multiple dimensions within this context.
        
        In the following two sections, we introduce new techniques for measuring reliability and demonstrate how to aggregate decisions across multiple criteria to generate an overall ranking using the BCJ framework.

\section{Measuring Reliability}
    \label{sec:reliability}

    Previously, Gray \textit{et al.} suggested that one can track the maximum entropy across all the possible pairs due to the Beta posterior distribution in each, and when it is \textit{sufficiently} low, one can stop selecting further pairs \cite{gray2023bayesian}. However, an entropy value can be difficult to interpret, and only makes sense as a relative measure, making it challenging to measure and communicate reliability, or to devise a stopping criterion for pair selection. 

\begin{figure}[t!]
        \centering
        \includegraphics[width=1\linewidth]{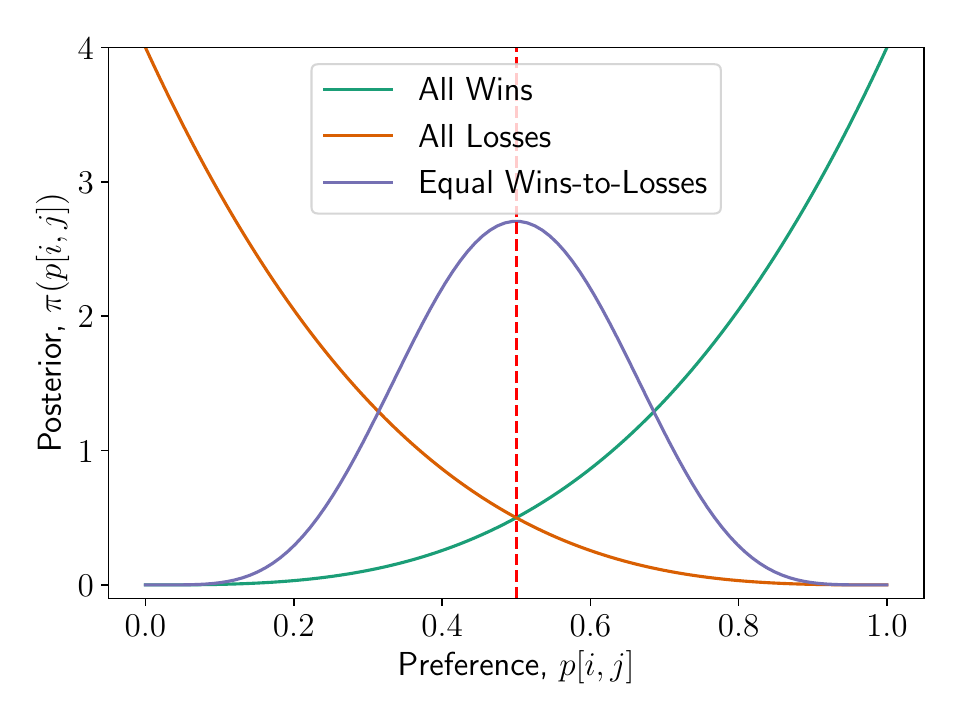}
        \caption{An illustration of the posteriors under different levels of agreements. When all ratings agree, on either all wins (shown in green), or all losses (shown in orange), for item $i$ compared to item $j$, the densities skew towards 1 or 0 respectively, with the corresponding most likely predicted outcome being close to 1 or 0. On the other hand, if we have the equal number of wins and losses, i.e. the highest level of disagreements between ratings, we get the purple density with the most likely outcome being 0.5 (depicted with the red dashed vertical line). Here, we assumed 4 comparisons have been made; with more comparisons, variance would reduce given the assumptions for outcomes. }
        \label{fig:pref_post_plot}
    \end{figure}

    One of the key feature of estimating posterior Beta distribution over the preference between two items is that it is directly encapsulating the level of agreement between the decisions that were made about a particular pair. This means when a pair truly divides the crowed (be it inter or intra rater), the probability Beta posterior distribution would have an expected value of 0.5, where 0 represents perfect agreement on an item losing and 1 represents the same item winning; see Figure \ref{fig:pref_post_plot} for an illustration of these possible cases.

    With this, we can formulate measures of reliability that diverges from the expected highest level of disagreement of 0.5. Given the most likely value of a Beta posterior is the mode, we can, firstly, define it to capture the divergence of the mode from 0.5. Noting that the direction of divergence does not matter, we define the mode agreement percentage (MAP) as follows:

    \begin{align}
        MAP(\alpha_{post}, \beta_{post}) = \frac{|m(\alpha_{post}, \beta_{post}) - 0.5|}{0.5}\times 100 \%,
    \end{align}

    where the mode $m(\alpha_{post}, \beta_{post}) = \frac{\alpha_{post}-1}{\alpha_{post} + \beta_{post} - 2}$ with $\alpha_{post}$ and $\beta_{post}$ are the posterior parameters for the Beta density over preference for a pair.

    While this provides an intuitive avenue to measure reliability, it does not appropriately incorporates the uncertainty from the paucity of comparison data per pair. To capture the uncertainty in a measure, we, therefore, propose to calculate the expected agreement percentage (EAP) as follows: 

    \begin{align}
        &EAP(\alpha_{post}, \beta_{post}) = \kappa \int_0^1 p^{\theta_{1}} \left(1 - p\right)^{\theta_{2}} \left|{p - 0.5}\right|\, dp\nonumber\\
        &= - \kappa \left[  \frac{0.5 \Gamma\left(\theta_{1} + 1\right) {{}_{2}F_{1}\left(\begin{matrix} - \theta_{2}, \theta_{1} + 1 \\ \theta_{1} + 2 \end{matrix}\middle| {1} \right)}}{\Gamma\left(\theta_{1} + 2\right)} \right.
        \nonumber \\&- \left. \frac{1.0 \Gamma\left(\theta_{1} + 2\right) {{}_{2}F_{1}\left(\begin{matrix} - \theta_{2}, \theta_{1} + 2 \\ \theta_{1} + 3 \end{matrix}\middle| {1} \right)}}{\Gamma\left(\theta_{1} + 3\right)} \right] 
        \nonumber \\&+ 2 \kappa \left(\frac{0.25 {0.5}^{\theta_{1}} \Gamma\left(\theta_{1} + 1\right) {{}_{2}F_{1}\left(\begin{matrix} - \theta_{2}, \theta_{1} + 1 \\ \theta_{1} + 2 \end{matrix}\middle| {0.5} \right)}}{\Gamma\left(\theta_{1} + 2\right)} 
        \right.\nonumber\\&
        \left.
        - \frac{0.25 {0.5}^{\theta_{1}} \Gamma\left(\theta_{1} + 2\right) {{}_{2}F_{1}\left(\begin{matrix} - \theta_{2}, \theta_{1} + 2 \\ \theta_{1} + 3 \end{matrix}\middle| {0.5} \right)}}{\Gamma\left(\theta_{1} + 3\right)}\right),
    \end{align}
    where, $\kappa = \frac{\Gamma(\alpha_{post}+\beta_{post})}{0.5 ~\Gamma(\alpha_{post})\Gamma(\beta_{post})}\times 100$, $\theta_{1} = \alpha_{post} - 1$, and $\theta_{2} = \beta_{post}-1$, with $\Gamma(\cdot)$ is the Gamma function and ${}_2F_1 (\cdot)$ is the Gaussian hypergeometric function. We validated this result through simulation.

    These formulations for MAP and EAP around 0.5 relate to percentiles over preferences. Specifically, the MAP (or EAP) metrics indicate how far the metric value is from the middle, on both sides, and thus inform us of the range beyond which we currently have the metric. We can calculate the lower bound of the range with \( l = 0.5 - \frac{0.5 ~ \text{MAP}}{100} \) and the upper bound of the range with \( u = 0.5 - \frac{0.5 ~ \text{MAP}}{100} \). For instance, a 50\% MAP means that the mode resides outside the range between \( l = 0.25 \) and \( u = 0.75 \). In terms of EAP, since this is integrated over the uncertainty in the density, a 50\% EAP would mean that there is enough volume to push the expected value of the agreement percentage beyond the range between \( l = 0.25 \) and \( u = 0.75 \). Hence, we can devise a stopping criterion based on the desired level of confidence, and thus enforce a range for this ``null space".

    Alternatively, the assignment owner can decide the lower and upper bounds of this ``null space" and then compute the threshold required for the minimum MAP or EAP before stopping further data collection. For example, if they wanted the width of the ``null space" to be $95\%$, they could define a range between \( l = 2.5\% \) and \( u = 97.5\% \), which would be equivalent to a threshold of 95\% on MAP or EAP (whichever they were tracking for this purpose).

\begin{figure}[t!]
            \centering
            \includegraphics[width=1\linewidth]{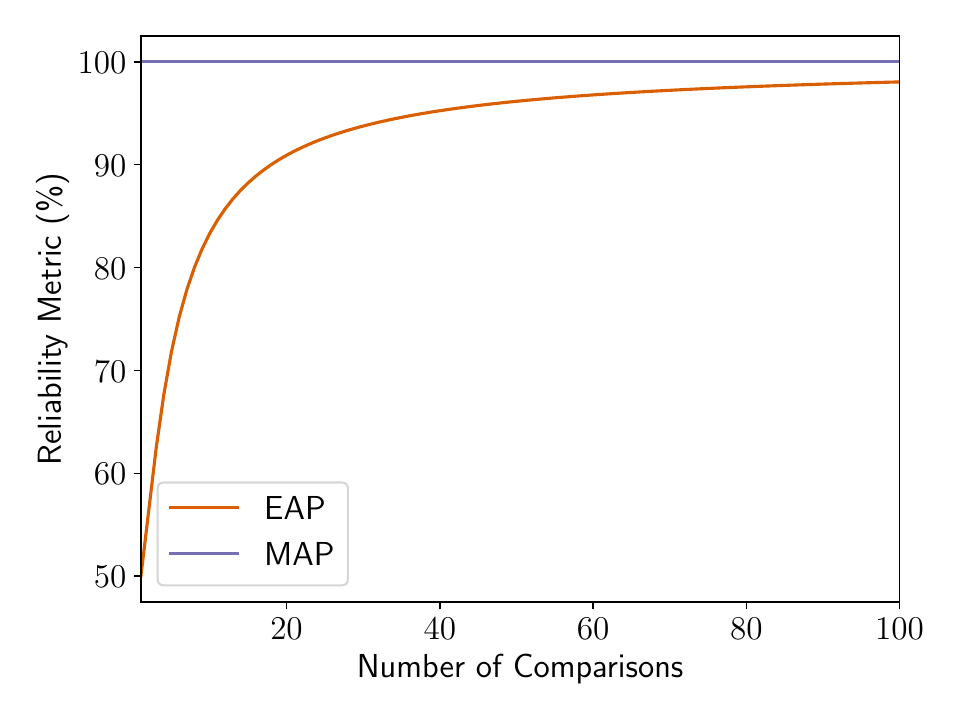}
            \caption{An illustration of EAP increasing slowly (shown in orange) as we observe an item winning at every comparison with another specific item to reflect the decreasing uncertainty over comparisons. Whereas MAP, shown in purple, is overoptimistic, and quickly gets to near 100\%, even with a few observed wins.}
            \label{fig:all_wins}
        \end{figure}

\begin{figure}[t!]
            \centering
            \includegraphics[width=1\linewidth]{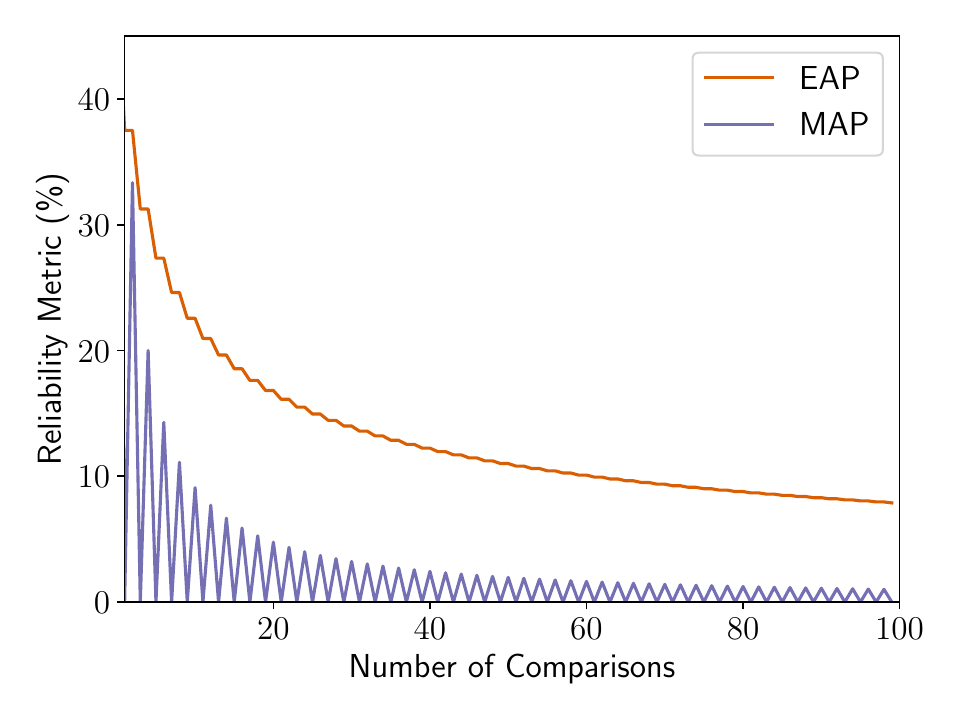}
            \caption{An example of EAP being more stable when there are conflicting information, with an item only winning every second comparison against a particular item. MAP fluctuates rapidly, but with sufficient data the overshoots are small (depicted in purple). }
            \label{fig:eq_wins}
        \end{figure}

    In terms of the choice of MAP or EAP, we noted that they both are useful in different ways. MAP provides an intuitive indication of where the mode is, but because it does not consider the level of existing uncertainty, it can be overly optimistic. On the other hand, EAP provides a more comprehensive metric of reliability that incorporates the amount of information at hand as we integrate over the uncertainty in the density. For example, consider a case when an item always wins in a pair. The mode would quickly shift towards the right even with a few wins, the mode would quickly shift towards the right, like the green line depicts in Figure \ref{fig:pref_post_plot}. However, the variance does not diminish so rapidly. Hence, the MAP will show a rather instantaneous shift towards 100\%, but EAP would only do so when there are numerous comparisons and all indicate wins for the item: see Figure \ref{fig:all_wins}. When the observations fluctuate between wins and losses, this instances shift in mode makes MAP fluctuate more acutely, especially when there is limited data; see Figure \ref{fig:eq_wins}.

    It should be noted that the decision to prefer one over the other in paired comparison may be made by the same individual at different times or different individuals (either synchronously or asynchronously), and the Bayesian machinery here would treat them the same way. Thus, both of these reliability metrics account for both inter and intra rater reliabilities depending on context of data collection. 

\section{Multi-Criteria Bayesian Comparative Judgement}
    \label{sec:mbcj}

   Suppose an assignment is evaluated against $D$ learning outcomes (LOs). The assignment owner specifies a weight vector $\bm{\lambda} = (\lambda_1, \dots, \lambda_D)^\top$, where each $\lambda_d \in [0, 1] \subset \mathbb{R}$ indicates the contribution of the $d$th LO to the overall mark, with the constraint that $\sum_{d=1}^D \lambda_d = 1$. Given a score $\gamma_{i,d}$ for item $i$ on LO $d$, the overall score for item $i$ is computed as: \begin{align} 
   \gamma_i = \sum_{d=1}^D \lambda_d \gamma_{i,d}. 
   \end{align}

    While this weighted aggregation is standard in rubric-based assessment, conventional CJ methods typically rely on holistic comparisons. This can obscure valuable insights into how an item performs across individual LOs. Although CJ provides fast and accurate overall rankings, assessors may find it difficult to revisit specific items and offer targeted feedback to learners on individual criteria.
    
    To adapt CJ for multi-criteria rubrics, we propose a framework in which decisions are made independently for each LO. This requires suitable methods for combining these decisions into a unified overall ranking. Additionally, we introduce a strategy for selecting the most informative pairwise comparison by considering uncertainty across all LO-specific evaluations. In the following section, we present novel methods to address these challenges.
    
    \subsection{Extension to Rank Generation}

    To generate a combined final rank for each item, we propose two approaches: one that aggregates LO-specific ranks, and another that merges LO-specific preference distributions, represented by Beta posterior densities.

    \subsubsection{Mixture of Component Ranks}
    \label{sec:rank-mix}

        For $d$th learning outcome LO$_d$, the rank distribution for item $i$ is denoted by $P(r_{i,d} = a)$, where $a \in [1, N] \subset \mathbb{N}$. Given a predefined weight vector $\mathbf{l}$, we can construct a mixture model to combine these LO-specific rank distributions as follows~\cite{lindsay1995mixture}: \begin{align} 
            P(r_i = a) = \sum_d \lambda_d P(r_{i,d} = a).
        \end{align} 
        The corresponding expected rank for item $i$ is then given by: 
        \begin{align} 
            \mathbb{E}[r_i] = \sum_d \lambda_d \mathbb{E}[r_{i,d}]. 
        \end{align}
    
\begin{figure}[t!]
            \centering
            \includegraphics[width=7.5cm]{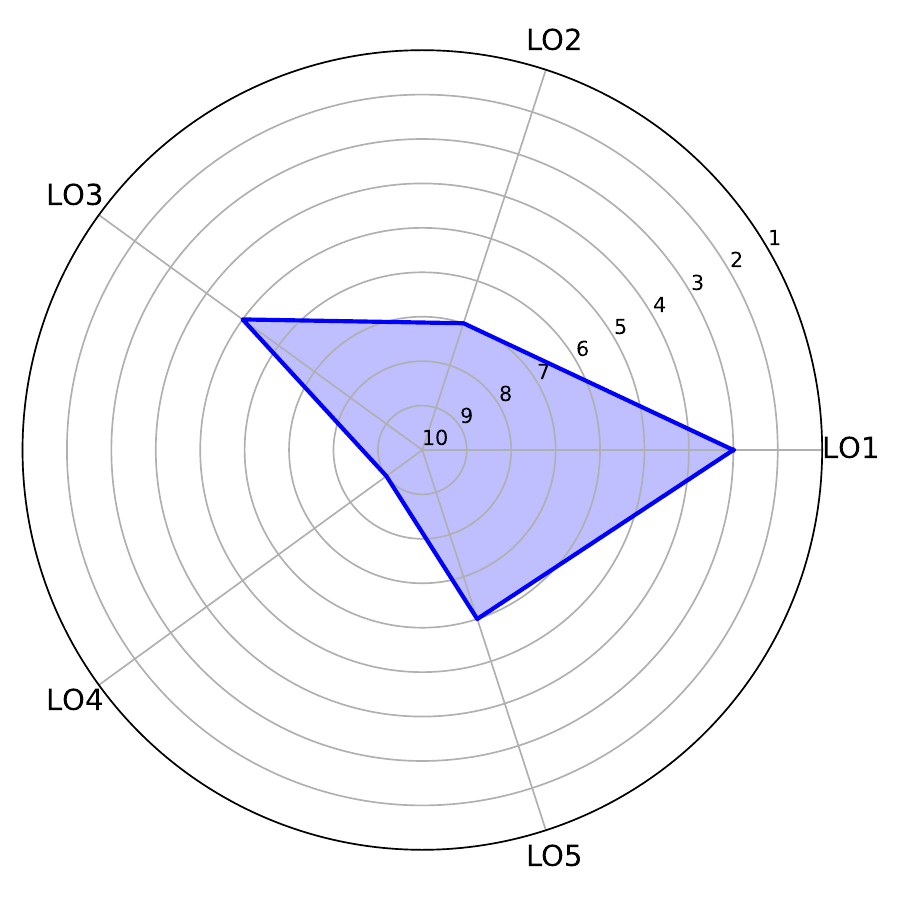}
            \caption{
            A radar plot depicting the $i$th item's $\mathbb{E}[r_{i,d}]$ performance across five LOs, enabling more transparency and detail on where this item performed well and where it did not.
    Therefore, it enables educators to identify areas where this candidate may need personalised intervention. Furthermore, it provides more insight than a traditional CJ rank would offer to the educator.}
            \label{fig:radar_plot}
        \end{figure}

        Figure \ref{fig:radar_plot} provides a visual illustration of the expected ranks $\mathbb{E}[r_{i,d}]$ for each LO. By combining these values using a weighted sum, we obtain the overall expected rank for item $i$.

        A central challenge in this approach lies in deriving the overall preference distribution $\pi(i \succ j)$ between items $i$ and $j$ that corresponds to these combined rank distributions. Without a likelihood-based model — such as the one employed in BTM — this estimation becomes non-trivial.

    \subsubsection{Mixture of Component Preferences}
        \label{sec:pref-mix}

\begin{figure}[t!]
            \centering
            \includegraphics[width=7.5cm]{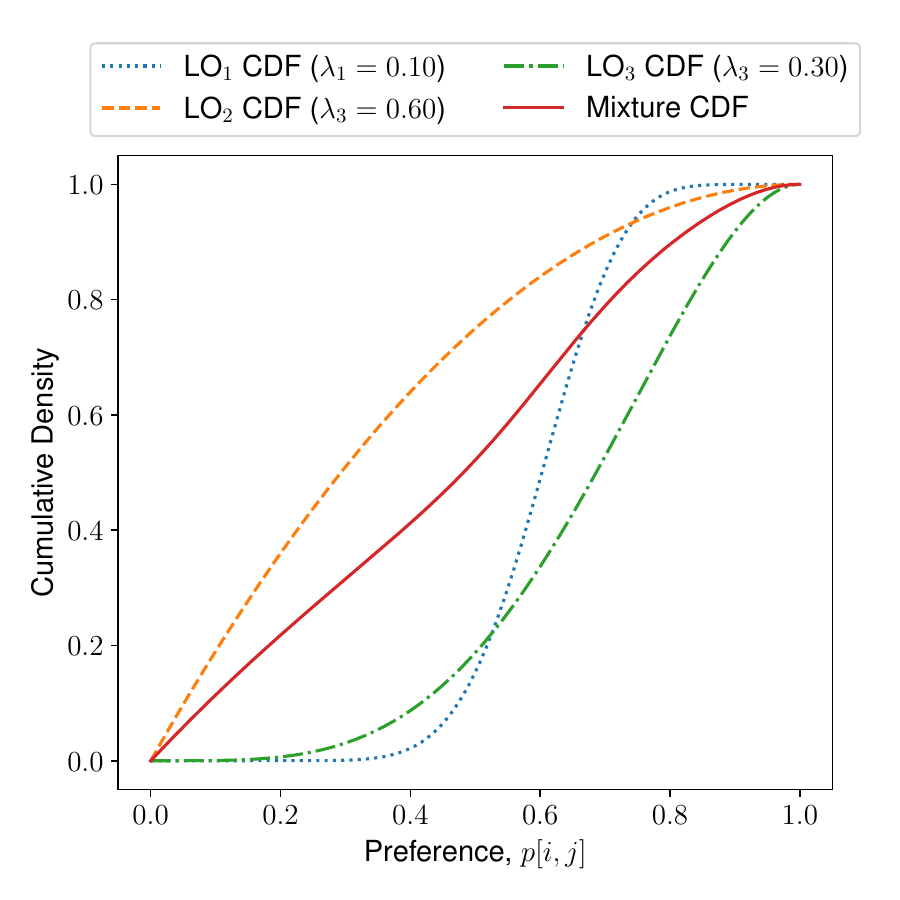}
            \caption{A visual illustration of how LO-specific preference distributions are combined using a weighted sum of their CDFs. In this example, three LOs are shown in blue, orange, and green, corresponding to the weight vector $\bm{\lambda} = (0.1, 0.6, 0.3)^\top$. The resulting mixture CDF, shown in red, is not a standard Beta distribution but effectively reflects the contributions of the individual components.}
            \label{fig:cdf_plot}
        \end{figure}

        Given the CDF of the preference distribution $\mathcal{F}_d(i \succ j)$ for the $d$th LO, the overall preference CDF for item $i$ over item $j$ can be expressed as a weighted mixture~\cite{lindsay1995mixture}: \begin{align} \mathcal{F}(i \succ j) = \sum_d \lambda_d \mathcal{F}_d(i \succ j), \end{align} where $\lambda_d$ denotes the contribution of LO$_d$ to the overall mark. We provide an illustration in Figure \ref{fig:cdf_plot} of such a mixture CDF.
    
        This formulation allows direct computation of the overall preference probability that item $i$ dominates item $j$, using Equation \eqref{eq:prob_pref}. It also enables derivation of the full probability distribution over ranks for each item. Methodologically, this approach is advantageous, as both preference and rank distributions can be efficiently obtained using existing mechanisms.
    
        However, exact calculation of overall preference and rank distributions becomes susceptible to combinatorial explosion when ranking a large number of items. To address this, a Monte Carlo (MC) sampling approach may be beneficial. For mixture of preference distributions, the $m$th sample $p_m$ can be drawn as follows: 
        \begin{align} p_m \sim \pi_q(i \succ j) ~ | ~ z_m \sim U(0,1) \wedge z_m \in \left[\sum_{d=0}^{q-1} \lambda_d, \sum_{d=0}^{q} \lambda_d\right], 
        \end{align} 
        where $z_m$ is the $m$th random number sampled from the uniform distribution $U(0, 1)$, $\pi_q(\cdot)$ is selected based on the value of $z_m$, and $l_0 = 0$. A win for item $i$ can be simulated by rounding the sampled proportion to the nearest integer, $x_m = \lfloor p_m \rceil$. The total number of wins and losses across all pairwise comparisons, aggregated over all preference distributions, can then be used to estimate rank distributions and compute the expected overall rank for each item.
    
    \subsection{Extension to Pair Selection}
        \label{sec:entropy_extension}

        Differential entropy generalises the classical concept of discrete entropy to continuous random variables, offering a measure of uncertainty associated with a probability distribution \citep{abramowitz1972stegun}. For the Beta distribution -- defined on a finite interval $[0,1]$ and parameterised by two shape parameters -- differential entropy reflects a measure of uncertainty over that interval, as discussed in Section \ref{bcj:pair-del}.
    
        In the multi-dimensional case, each preference distribution $\pi_d(i \succ j)$ between items $i$ and $j$ for LO$_d$ is modelled as a Beta distribution $\mathcal{B}(\alpha_{\text{post}}, \beta_{\text{post}})$. Assuming independence across LOs, the total entropy across all $D$ LOs is then computed as \cite{korn2000mathematical}: 
        \begin{align} 
            H(\pi(i \succ j)) = \sum_d H(\pi_d(i \succ j)). 
        \end{align} 
        This enables selection of the most informative pairwise comparison by identifying the pair with the highest total entropy: 
        \begin{align} 
            (i, j) \gets \arg \max_{(i,j) \wedge i \neq j} H(\pi(i \succ j)). 
        \end{align}

\section{Experimental Setup}
    \label{sec:model,data,metrics}

    In this paper, we aim to experimentally investigate the following questions:
    \begin{enumerate}[label=(Q\arabic*)]
        \item How can reliability be assessed at a granular level, and how does this inform targeted moderation?\\
        \textit{(Results discussed in Section~\ref{sec:reliability-test})}
        
        \item Which combinations of ranking and selection strategies perform best in multi-criteria CJ when weights are fixed?\\
        \textit{(Results discussed in Section~\ref{sec:best-start})}
        
        \item Which approaches remain robust under varying weight configurations?\\
        \textit{(Results discussed in Section~\ref{sec:robust-test})}
    \end{enumerate}

    This section begins by outlining the strategy variants included in our comparison, followed by an overview of the real-world datasets used. We then describe the simulation process for modelling assessor decision-making and conclude with a summary of the performance evaluation methodology.

    \subsection{Strategies Under Scrutiny}
    \label{sec:strategies}

        We explore six strategy combinations formed by pairing two ranking methods -- Ranking Mixture Model (Section \ref{sec:rank-mix}) and Preference Mixture Model (Section \ref{sec:pref-mix}) -- with three pair selection techniques: entropy-based (Section \ref{sec:entropy_extension}), random, and NRP. These combinations are evaluated against the baseline BCJ ranking method with entropy-driven selection, as proposed by Gray \textit{et al.}.

    \subsection{Datasets}
    \label{sec:datasets}

        In this paper, we use two datasets: the DREsS dataset \citep{yoo2024dress} and a curated dataset from an undergraduate assignment at a British university (BU). Both datasets contain item identifiers and corresponding absolute marks assigned by human assessors. These marks serve as ground truth rankings and are also essential for simulating decision-making processes, as discussed in Section \ref{sec:sim_dec}. Importantly, the datasets deploy different weighting, and as such they offer a valuable opportunity to evaluate the applicability and efficacy of the active ranking strategies proposed in this paper.
    
        The DREsS dataset \citep{yoo2024dress} is drawn from a real classroom setting and includes approximately 1,700 essays written by undergraduate students who are English as a Foreign Language (EFL) learners. It consists of three sub-datasets: $DREsS_{New}$, $DREsS_{Std.}$, and $DREsS_{CASE}$, each designed for different aspects of automated essay scoring (AES). For our study, we focus solely on $DREsS_{New}$, as it best represents authentic EFL writing scenarios.
        
        Essays in this dataset are evaluated using a rubric with three equally weighted criteria \citep{yoo2024dress}: Content (relevance and depth of the subject matter), Organisation (structure, coherence, and logical flow), and Language (grammar, vocabulary, and overall language use). Each criterion is scored out of 5 by trained English education experts, resulting in a total score out of 15.
        
        The BU dataset also originates from a real classroom and includes 69 assessment records from a first-year undergraduate module. Students were asked to choose from several scenarios and develop a web page based on their selection, demonstrating specific skills. For consistency, we selected 38 samples based on the same scenario. These were assessed on three criteria: implementation quality of core components, fulfilment of additional brief requirements, and documentation quality. Each criterion is scored out of 100, with respective weightings of 50\%, 25\%, and 25\% towards the final mark.

        To run the experiments, we select subsamples of size $N \in \{5, 10, 15, 20, 25\}$ from each dataset to generate the ground truth target ranks using the absolute marks. We use stratified sampling \cite{neyman1992two} to ensure good coverage across the range of marks in the subsets. We then choose a multiplier $K \in \{5, 10, 20, 30\}$ to determine the overall budget, i.e., $N \times K$, representing the number of comparisons performed under each strategy.
    
    \subsection{Automated Decision Simulation}
        \label{sec:sim_dec}
        
        In the absence of human decision-makers — although we plan to conduct real experiments involving human participants in the future — it is necessary to simulate the decision-making process in a plausible and principled way. To achieve this, we adopt the method proposed by Gray \textit{et al.}, where each item's ground truth mark is treated as the mean of a Normal distribution. The standard deviation reflects typical tolerance levels observed in marker disagreements.
        
        For the DREsS dataset, where no formal guidance on tolerance is available, we set the standard deviation to $\sigma = 0.5$. This corresponds to a discrepancy of $\pm 2\sigma = \pm 20\%$ in absolute marks, encompassing approximately 95\% of the probability mass under a Normal distribution. This represents a relatively relaxed marking scenario. In contrast, for the BU dataset, we follow established tolerance guidelines and set the standard deviation to $\pm 3\%$, resulting in an acceptable discrepancy of around $\pm 6\%$ with 95\% probability. These two contrasting scenarios allow us to explore the impact of uncertainty in the marking process — from high variability in DREsS to a more stringent and consistent marking regime in BU.
        
        Under this setup, we simulate pairwise comparisons by sampling from two Normal distributions. Each item's score is drawn randomly, and the item with the higher score is considered the winner. Formally, for items $i$ and $j$, we sample $x_i \sim \mathcal{N}(\mu_i, \sigma_i)$ and $x_j \sim \mathcal{N}(\mu_j, \sigma_j)$, where $\mathcal{N}(\mu, \sigma)$ denotes a Normal distribution with mean $\mu$ and standard deviation $\sigma$. The simulated winner is determined as follows:
        
        \begin{align}
        w_{i,j} =
        \begin{cases}
        1 & \text{if } x_i \geq x_j \\
        0 & \text{otherwise}
        \end{cases}
        \end{align}

    \subsection{Performance Metrics}

        Given the ground truth target rank, we measure performance using the normalised Kendall's $\tau$ rank distance, which quantifies the difference between two ranking lists: the ground truth and the rank produced by the strategy under evaluation at a given point in time. The metric is calculated by counting the number of pairwise discrepancies between the two lists. A greater distance indicates greater disagreement~\citep{kendall1938new, fagin2003comparing}. The normalised distance ranges from 0 (perfect agreement) to 1 (complete disagreement). For example, a distance of $0.03$ implies that only $3\%$ of the pairwise orderings differ. In this paper, we record the $\tau$ distance after each paired comparison to track how well each method converges to the target rank.
        
        Each strategy is run independently for a given budget over 50 trials, and the final $\tau$ scores are recorded. These scores are then statistically compared. We use the one-tailed Wilcoxon rank-sum test (also known as the Mann–Whitney U test), a nonparametric test used to determine whether one group performs significantly better than another across two independent samples~\citep{macfarland2016introduction}. This test is often considered the nonparametric equivalent of the Student’s t-test for independent samples.
        
        Since we compare $7$ strategies for each budget level, we apply a Bonferroni correction~\citep{dunn1961multiple} to adjust for multiple comparisons. The significance level is modified as follows: $\alpha_{sig} \gets \frac{0.05}{S}$, where $S$ is the number of strategies being compared.
        
        In the next section, we discuss our findings from the experimental studies.

\section{Results and Discussion}
    \label{sec:experiment_discussion}

    \subsection{Assessing Reliability and Integrating Principal Marker Interventions}
    \label{sec:reliability-test}

        In this section, we begin our investigation into assessing reliability using single-criterion BCJ. For an arbitrary instance of the DREsS dataset with $N=10$ and a budget multiplier $K=10$, we run BCJ for $N \times K = 100$ simulated pairwise comparisons, driven by entropy-based selection. We record the final $\tau$ score with respect to the ground truth, along with the MAP and EAP scores for each pairwise comparison.

\begin{figure*}[ht!]
            \centering
            \includegraphics[width=\textwidth]{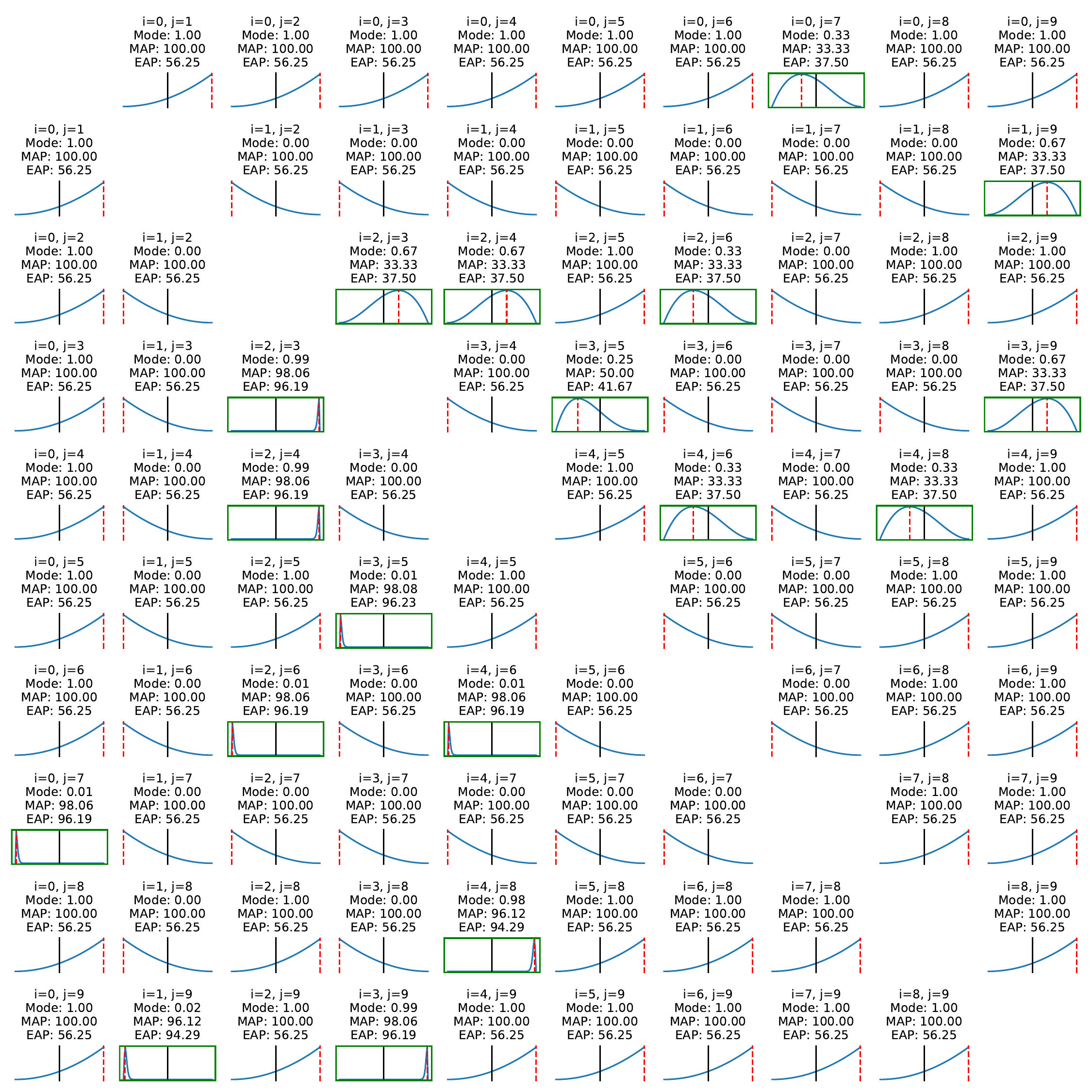}
            \caption{MAP, and EAP scores for each pairwise comparison in the DREsS dataset ($N=10$, $K=10$). Comparisons with EAP scores below 50\% were flagged and reviewed by a PM to identify items causing disagreement (shown within green boxes). The PM then selected the winner and we biased the respective preference distributions accordingly. The upper triangle displays the original decisions prior to intervention, while the lower triangle reflects the updated outcomes after moderation.}
            \label{fig:eap_map_lmod}
        \end{figure*}

        Unlike a single SSR score, the proposed MAP and EAP metrics offer a more nuanced perspective on assessor agreement. These metrics enable the identification of specific item pairs that contribute to disagreement, providing actionable insights that a single, aggregate reliability score cannot capture. By using MAP and EAP, we gain a clearer understanding of uncertainty at the pairwise level (see the upper triangle of Figure \ref{fig:eap_map_lmod} for a visual illustration). Importantly, this analysis pertains to reliability -- whereas ranking accuracy is determined solely by the Bayesian estimation of rank distributions, which is theoretically optimal given the available data.

        To showcase the practical value of the EAP metric, we simulated an intervention by a principal moderator (PM). By flagging item pairs with low agreement (EAP < 50\%), the PM could concentrate on the most contentious comparisons and make a judgement on which item should be considered superior. The chosen item in each pair was then assigned an artificial win count of $1000$, indicating strong confidence in the decision and effectively removing it from future selection via entropy-based methods. 
        
        Figure \ref{fig:eap_map_lmod} illustrates this process: the upper triangle highlights low-agreement pairs (EAP < 50\%) with green boxes, while the lower triangle displays the updated EAP scores and distributions following the PM’s intervention.
    
        The ground-truth target ranking for the items was:
        \begin{align}
            \langle 7, 0, 6, 2, 5, 4, 3, 8, 9, 1 \rangle\nonumber,
        \end{align}
        while the estimated ranking prior to intervention was:
        \begin{align}
            \langle 7, 0, 6, 2, 4, 5, 8, 3, 1, 9 \rangle\nonumber.
        \end{align}
        
        This reveals three misordered pairs: $\langle 5, 4 \rangle$, $\langle 3, 8 \rangle$, and $\langle 9, 1 \rangle$. Following the intervention, two of these — $\langle 5, 4 \rangle$ and $\langle 9, 1 \rangle$ — were corrected. The pair $\langle 8, 3 \rangle$ remained unresolved, as item 3 had only been compared with items 2, 5, and 9, and was consistently judged superior. Item 8 was also deemed better than those, leaving no new information on direct comparison between items 3 and 8 to inform their relative ranking. Nevertheless, this targeted intervention improved ranking accuracy, reducing the $\tau$ score from 0.07 to 0.04.
            
        Importantly, we do not account for probabilistic transitivity due to the independence assumption in BCJ, which aligns with findings in human decision-making behaviour \cite{tversky1969substitutability}. While incorporating inter-item dependencies could enhance accuracy, it would significantly complicate the probabilistic modelling.

    \subsection{Identifying the Best Strategy}
        \label{sec:best-start}

        \begin{figure*}[]
            \centering
            \begin{tabular}{c c c c}
                \hline
                 & Entropy & Random & No Repeating Pairs \\
                \hline
                \raisebox{50pt}[0pt][0pt]{\rotatebox[origin=c]{90}{MCP}}& \includegraphics[width=5cm]{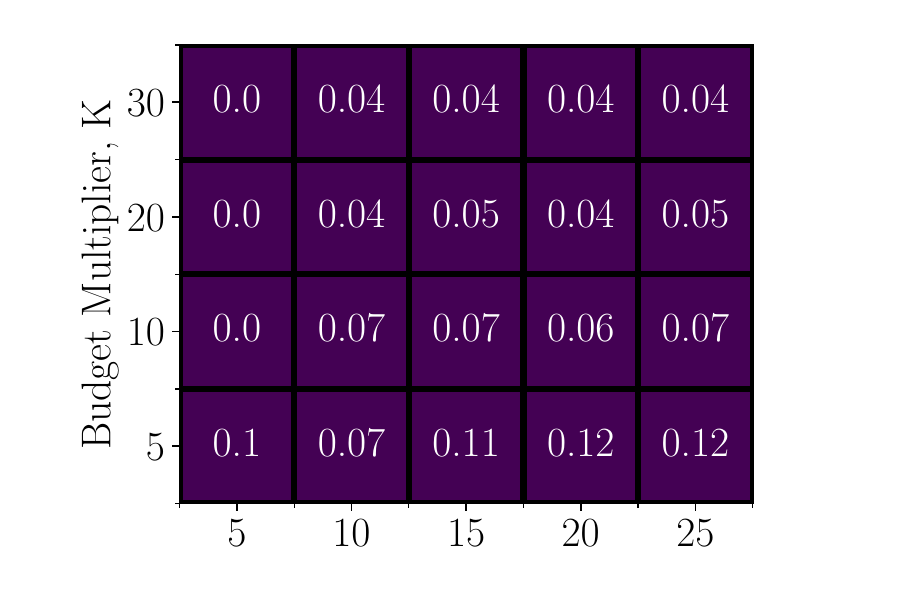} & \includegraphics[width=5cm]{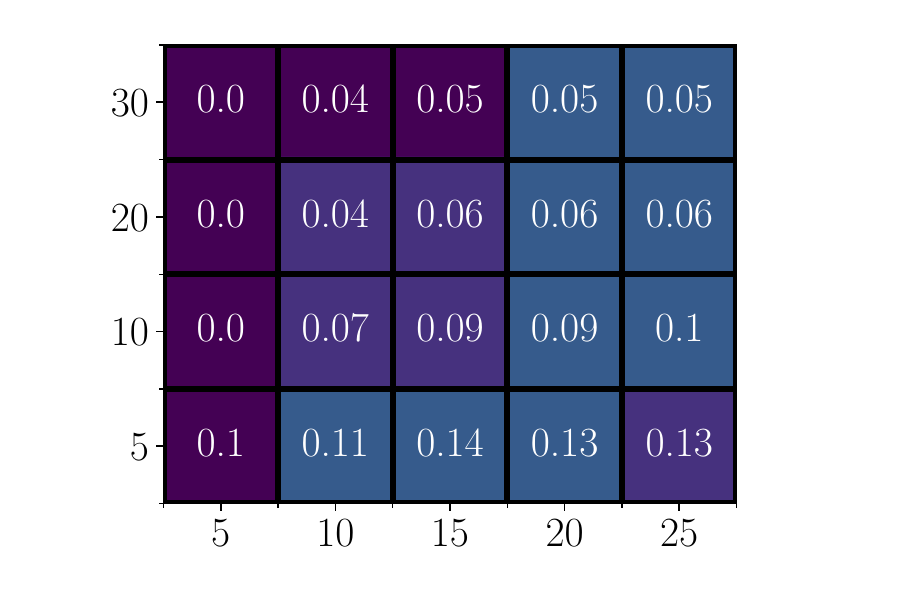} & \includegraphics[width=5cm]{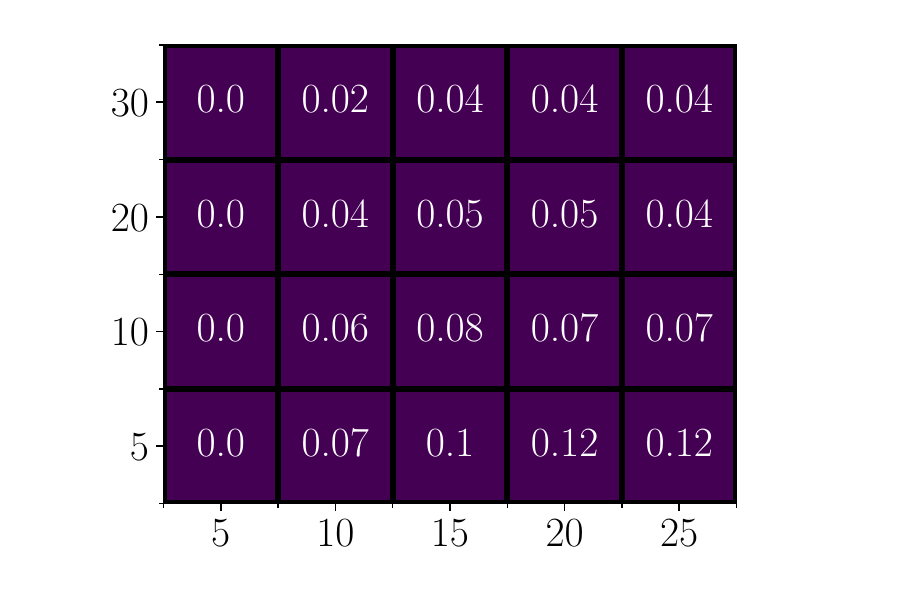} \\
                \hline
                \raisebox{50pt}[0pt][0pt]{\rotatebox[origin=c]{90}{MCR}} 
                & \includegraphics[width=5.5cm]{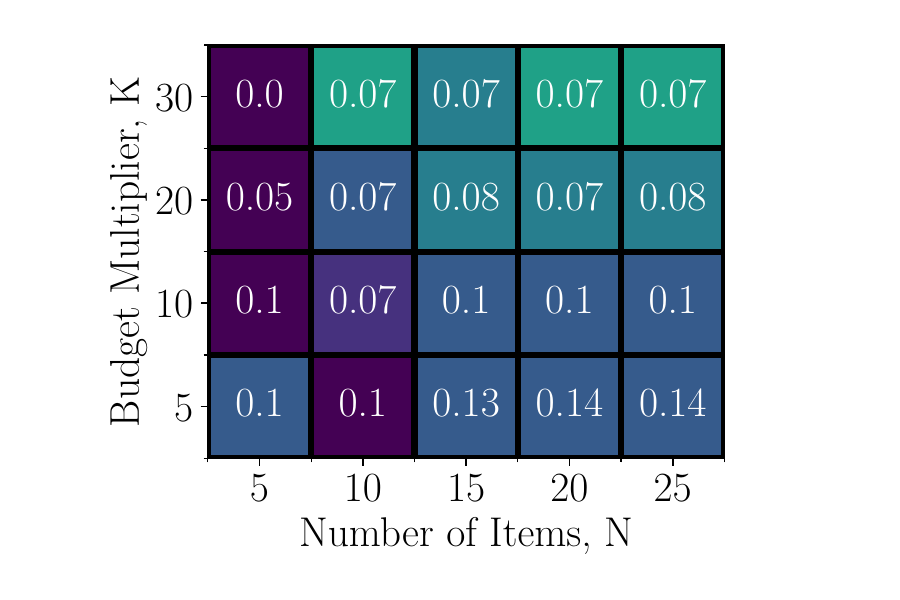} & \includegraphics[width=5.5cm]{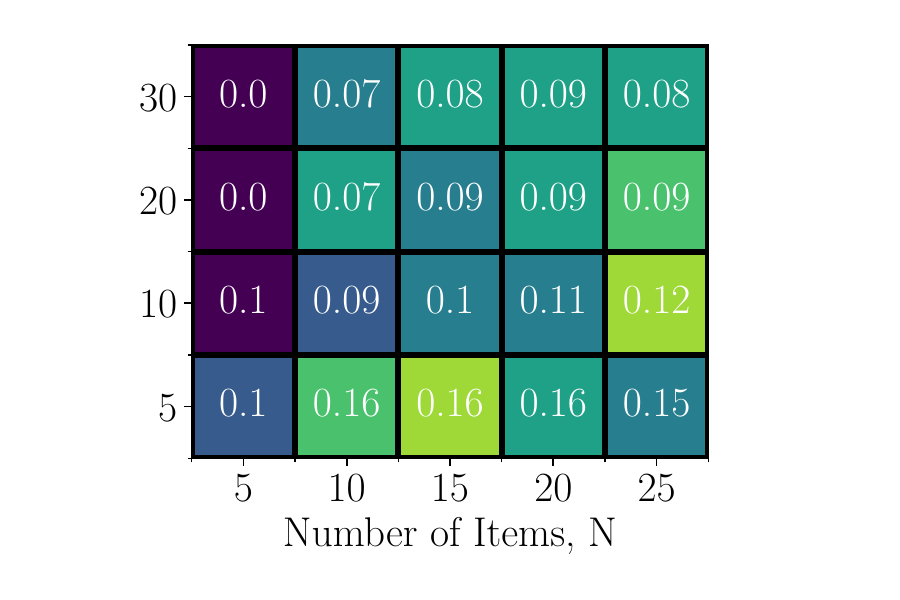} & \includegraphics[width=5.5cm]{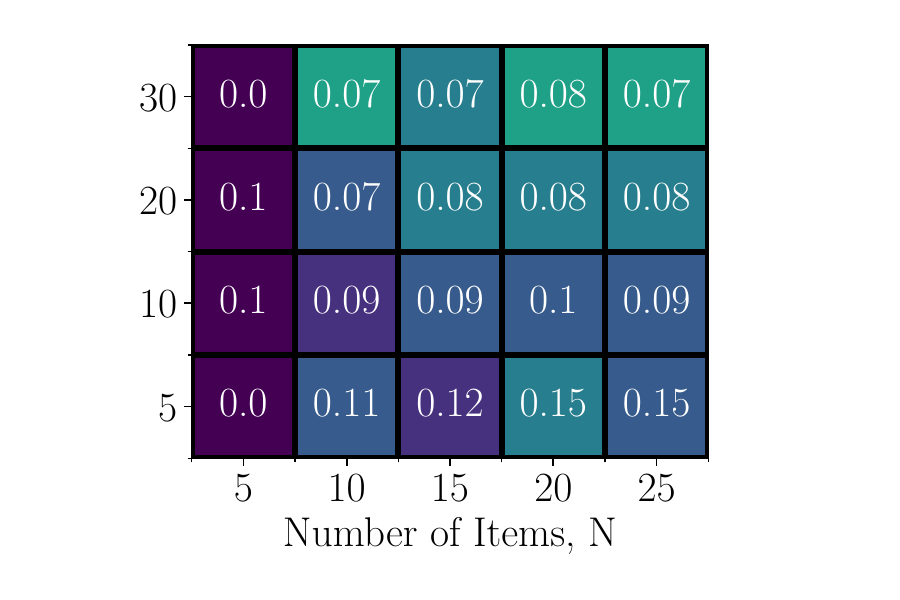} \\
                \hline
                \multicolumn{4}{c}{\includegraphics[width=6.5cm]{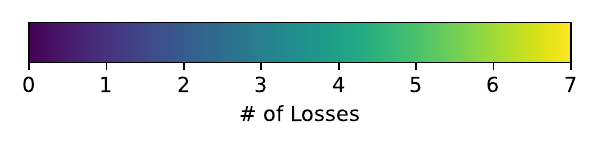}} \\
            \end{tabular}   
            \caption{
             Statistical comparison of results from the Wilcoxon rank-sum test on the DREsS dataset for multi-criteria strategies based on the mixture of component ranks (MCR) and the mixture of component preferences (MCP). Each strategy in a panel is identified by the row and column labels. Each cell is coloured according to the number of items (horizontal axis) and the budget multiplier $K$ (vertical axis). The colour of each cell reflects how often a particular strategy was outperformed by another competing strategy: darker colour indicates stronger performance (fewer losses), while lighter colour indicates weaker performance (more losses). The number shown in white within each cell represents the respective median performance. An MCP based strategy incorporating the entropy and NRP pair selection methods demonstrate the best overall performance across the experiments with this dataset, with the entropy method a close second.}
            \label{fig:wcrs_results_d}
        \end{figure*}

        For every strategy -- defined as a combination of ranking and pair selection methods -- we conduct $50$ repeated runs for a given budget $N \times K$, recording the final $\tau$ scores with respect to the ground truth. These results allow us to compare strategies using the Wilcoxon rank-sum test and identify the statistically superior approaches.
        
\begin{figure}
            \centering
            \includegraphics[width=7cm]{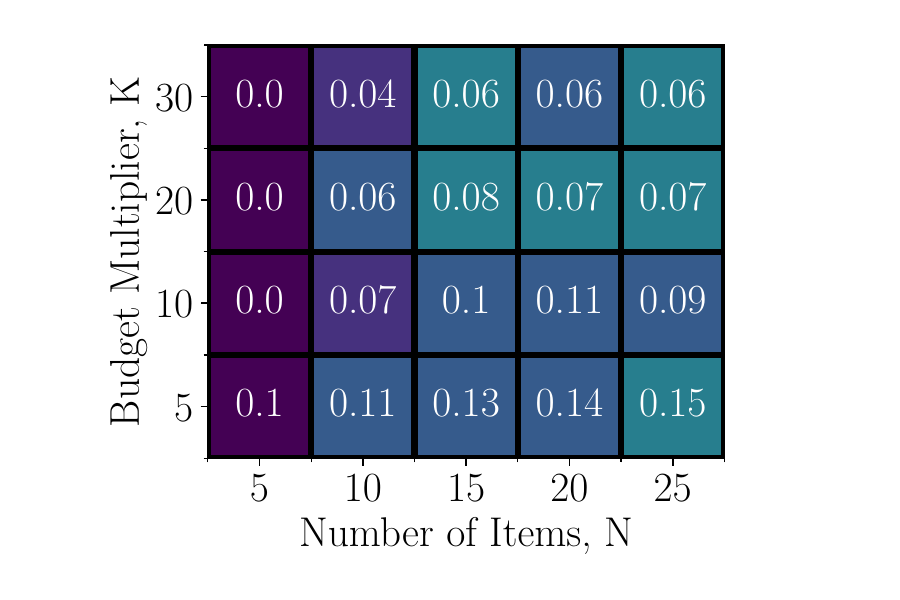}
            \caption{An illustration of the statistical comparison results for single-criteria (holistic BCJ with entropy-based pair selection) versus multi-criteria strategies. The colour of each cell represents how many times a given strategy was outperformed by others. Each cell displays the corresponding median performance in white. For $N = 5$, the BCJ strategy performs comparably to the best-performing strategy. However, for $N \geq 10$, there is at least one other strategy that consistently outperforms BCJ.}
            \label{fig:bcj_dress}
        \end{figure}

        In Figure~\ref{fig:wcrs_results_d}, we illustrate the results for the DREsS dataset. Among the multi-criteria strategies, MCP with entropy and NRP consistently outperforms the others, as it is never beaten by any competing strategy. It should be noted that MCR, regardless of the pair selection method, performs reasonably well beyond $N = 5$, with the weakest performance observed for MCR using random pair selection. For the single-criteria strategy—BCJ with entropy—shown in Figure~\ref{fig:bcj_dress}, we observe that it is outperformed by at least one other strategy when $N \geq 10$.

\begin{figure*}[]
            \centering
            \begin{tabular}{c c c c}
                \hline
                 & Entropy & Random & No Repeating \\
                \hline
                \raisebox{50pt}[0pt][0pt]{\rotatebox[origin=c]{90}{MCP}} & \includegraphics[width=5cm]{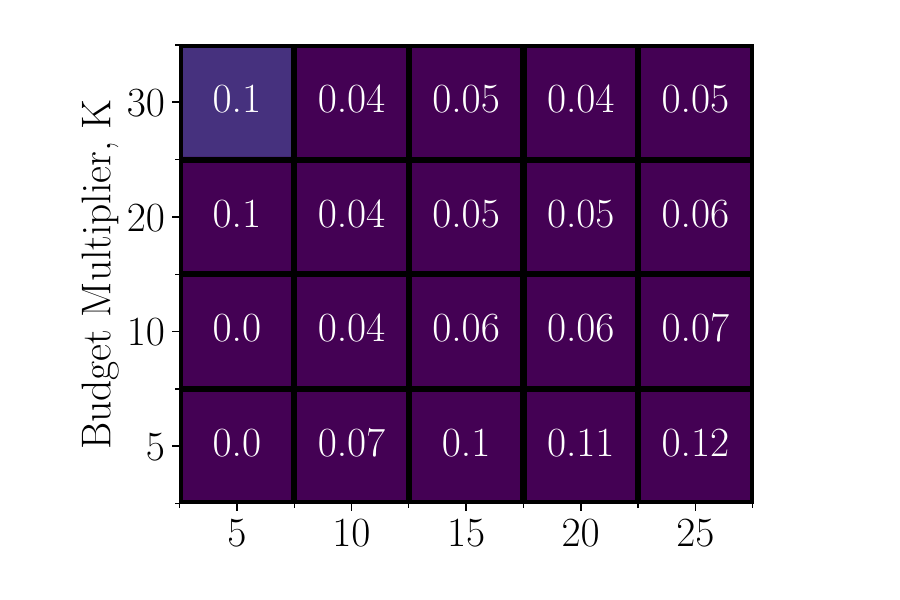} & \includegraphics[width=5cm]{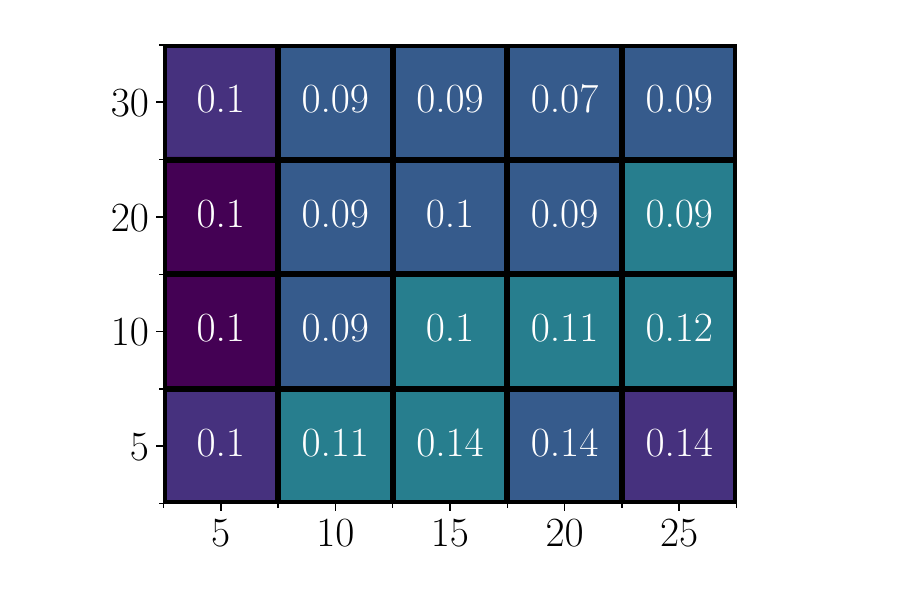} & \includegraphics[width=5cm]{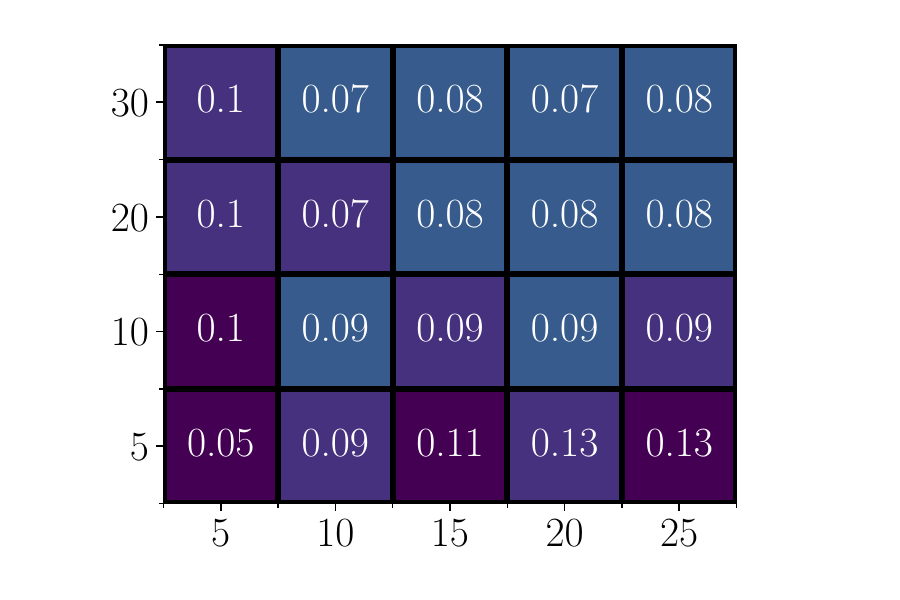} \\
                \hline
                \raisebox{50pt}[0pt][0pt]{\rotatebox[origin=c]{90}{MCR}} & \includegraphics[width=5.5cm]{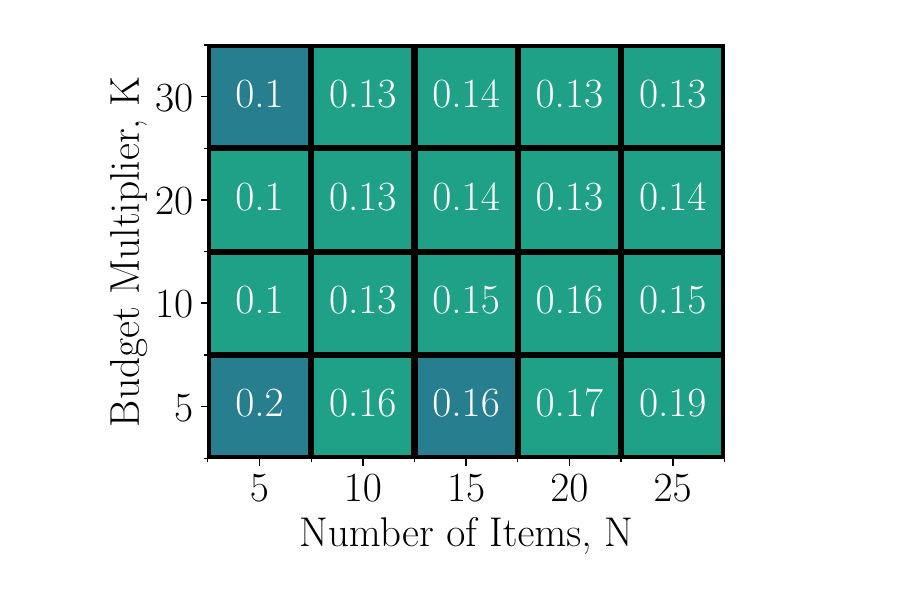} & \includegraphics[width=5.5cm]{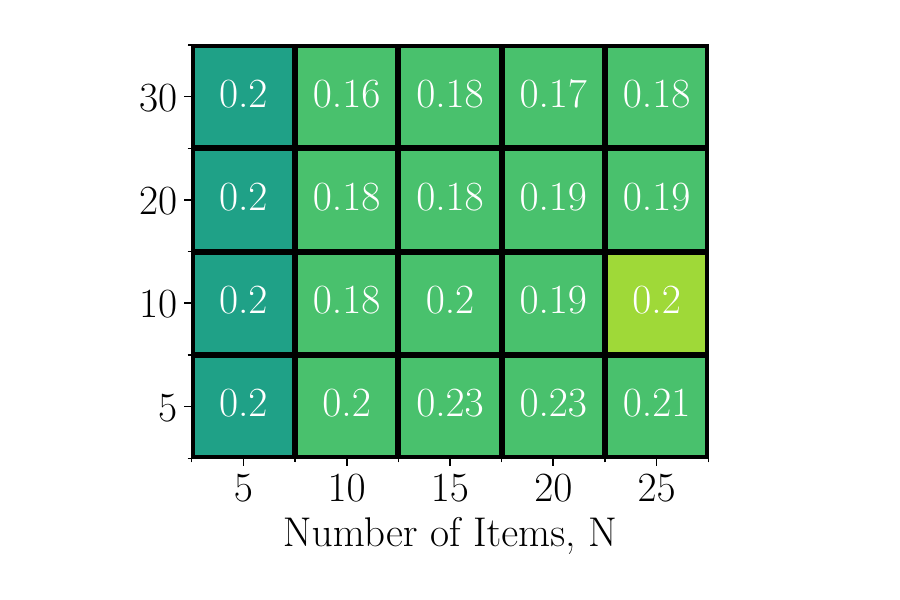} & \includegraphics[width=5.5cm]{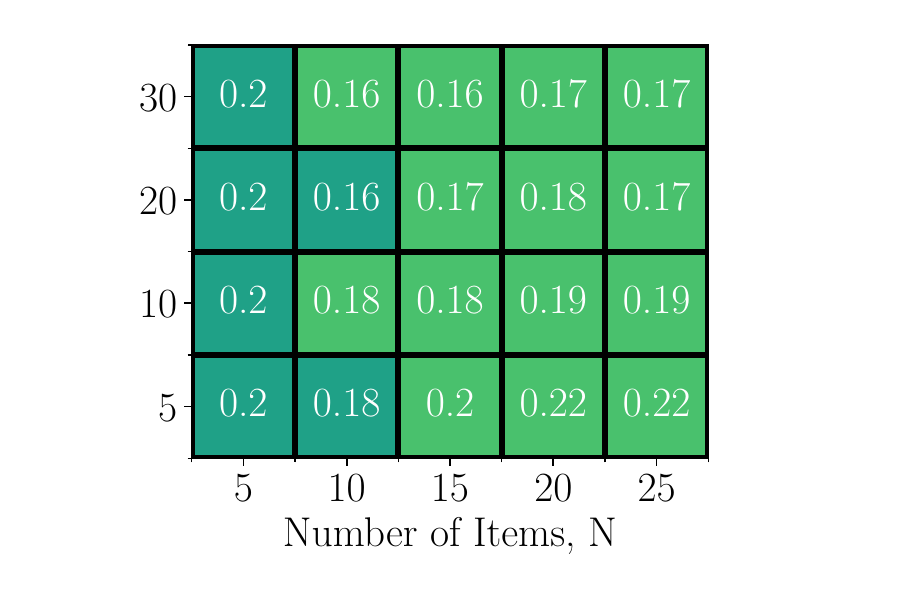} \\
                \multicolumn{4}{c}{\includegraphics[width=6.5cm]{cplots/COLOURBAR_ONLY.pdf}} \\
            \end{tabular}   
            \caption{
            An illustration of the statistical comparison results from the Wilcoxon rank-sum test for multi-criteria strategies based on the mixture of component ranks (MCR) and the mixture of component preferences (MCP) on the BU dataset. The plots show the number of times each combination of ranking method and pair selection method was the best, or equivalent to the best. The darkest colour indicates that the strategy was not beaten by any other method for that configuration, including comparisons against the single-criteria BCJ strategy using the standard entropy-based pair selection method. The white number in each cell indicates the median performance for that category. The MCP strategy demonstrates the strongest overall performance across the experiments, being beaten only once at the $N = 5$, $K = 30$ configuration.}
            \label{fig:wcrs_results_bsu}
        \end{figure*}

\begin{figure}
            \centering
            \includegraphics[width=7cm]{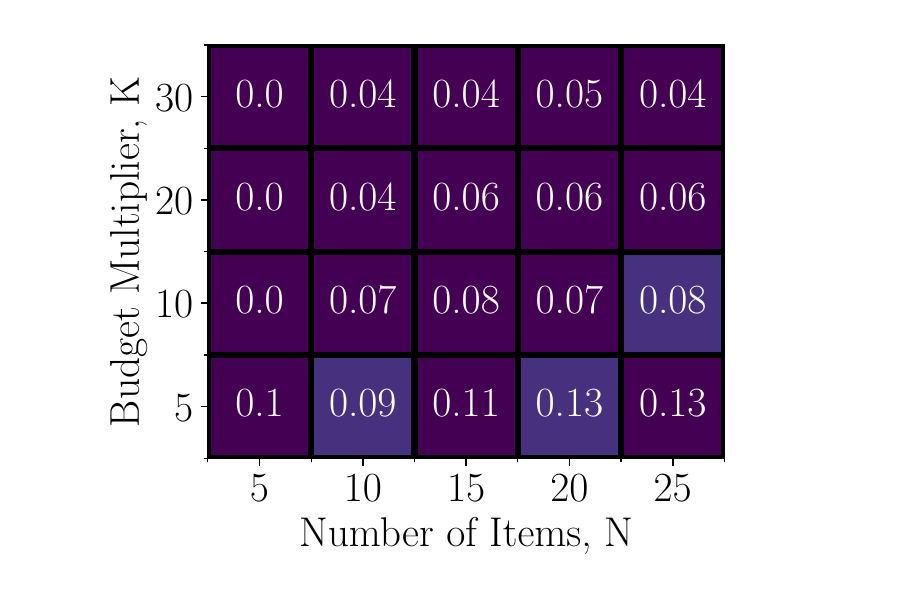}
            \caption{An illustration of the statistical comparison of results of the Wilcoxon rank-sum test for the level 4 undergrad dataset of the BCJ and entropy picking methods against the multi-criteria BCJ and other picking methods. We can see for this dataset apart from $N=10$ and $N=20$ for the $K$ value $5$, this approach was not dominated by any of the other combinations.}
            \label{fig:wcrs_bcj_results_bsu}
        \end{figure}

        For the BU dataset, which features tighter marking tolerance, we again observe that MCP is the superior approach among the multi-criteria strategies. In this case, the entropy-based pair selection method performs best, being beaten only once at $N = 5$ and $K = 30$ (see Figure~\ref{fig:wcrs_results_bsu}). In this instance, the single-criteria BCJ strategy also performs well (see Figure~\ref{fig:wcrs_bcj_results_bsu}), and is only outperformed for $N \in \{10, 20\}$ with $K = 5$, presumably due to limited preference data available at lower budget levels.

        Overall, we observe that the strategy combining MCP ranking with entropy-based pair selection performs best across both datasets. While the NRP method paired with MCP also performs well on the DREsS dataset, we attribute this to the greater pair-wise uncertainty due to the tolerance level associated with that dataset. Typically, entropy selects the most informative pair. When the preference differences between items are clear (i.e. more certain), those pairs are unlikely to be revisited frequently. This creates a distinction between entropy-based pair selection, which targets the most informative comparisons, and the NRP method, which iteratively revisits all pairs in a round-robin fashion. However, in cases of high uncertainty, both entropy and NRP tend to behave similarly, as more frequent revisiting of uncertain pairs becomes necessary.

        In terms of the single-criteria BCJ strategy, it generally performs well, particularly when uncertainty in overall preferences is low. However, by design, it lacks the richness provided by LO-specific preference information, which may limit its effectiveness in real-world assessment scenarios; especially from a feedback perspective.

    \subsection{Robustness to Varying Weight Configurations}
    \label{sec:robust-test}

    \begin{figure}
        \centering
        \includegraphics[width=\linewidth]{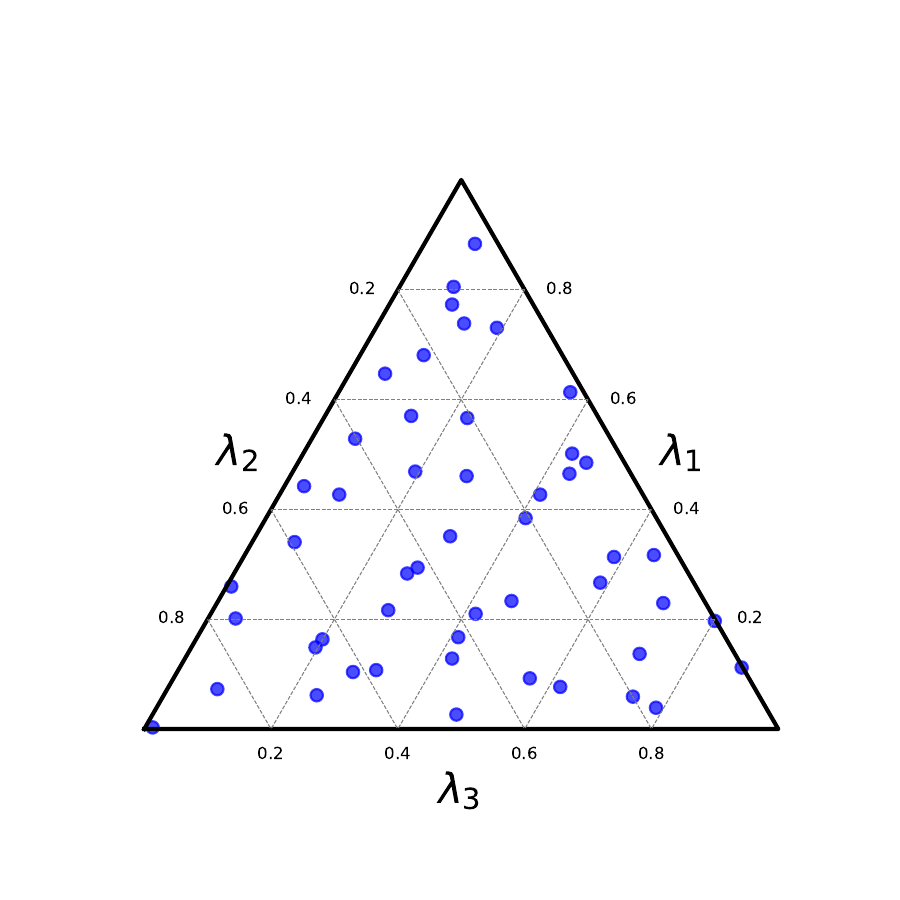}
        \caption{An illustration of the QMC weights transformed onto a simplex which is used for testing the robustness of the approaches.}
        \label{fig:simplex_weights}
    \end{figure}

    Both datasets come with predefined weights (as discussed in Section~\ref{sec:datasets}). Accordingly, we define a random performance vector $\tau(s ~|~ \bm{\lambda})$, where $s$ represents a strategy and $\tau (\cdot | \bm{\lambda})$ is the performance metric for a specified weight vector $\bm{\lambda}$. While we observe clear benefits from the MCP strategy with entropy-based pair selection, a natural question arises: what happens if we vary the weights? Would the conclusions remain the same?

    Addressing this question requires marginalising the effect of weights, i.e., estimating $\int \tau(s ~|~ \bm{\lambda}) ~d\bm{\lambda}$. In the absence of an analytical expression, this integral must be approximated. A standard MC method may not be ideal, as it requires thousands of samples, and for each sample, a simulation must be run across all budget configurations $N \times K$ -- a process that could take months or even years to complete.
    
    Instead, we estimate this using a Quasi-Monte Carlo (QMC) method with the Halton sequence~\citep{morokoff1995quasi}. QMC improves upon MC by replacing random sampling with low-discrepancy sequences, such as Halton, which are designed to cover the space more uniformly. These sequences minimise gaps and overlaps, leading to more accurate approximations with fewer samples, without sacrificing much in terms of accuracy. 

    To satisfy the constraint $\sum_d \lambda_d = 1$, we followed the method proposed by Smith \textit{et al.} \cite{smith2004sampling}. This technique generates sequences in $D-1$ dimensions, appends a value of 1, sorts the resulting list, and then computes the differences between adjacent values to derive the weights $\lambda_d$. This transformation effectively maps the original $D-1$ dimensional sequence onto a simplex — a geometric structure in $D-1$ dimensions where each point is $D$-dimensional and inherently satisfies the required weighted sum condition. See Figure \ref{fig:simplex_weights} for an illustration of the weights.

\begin{figure*}[]
            \centering
            \begin{tabular}{c c c c}
                \hline
                 & Entropy & Random & No Repeating Pairs \\
                \hline
                \raisebox{50pt}[0pt][0pt]{\rotatebox[origin=c]{90}{MCP}} & \includegraphics[width=5cm]{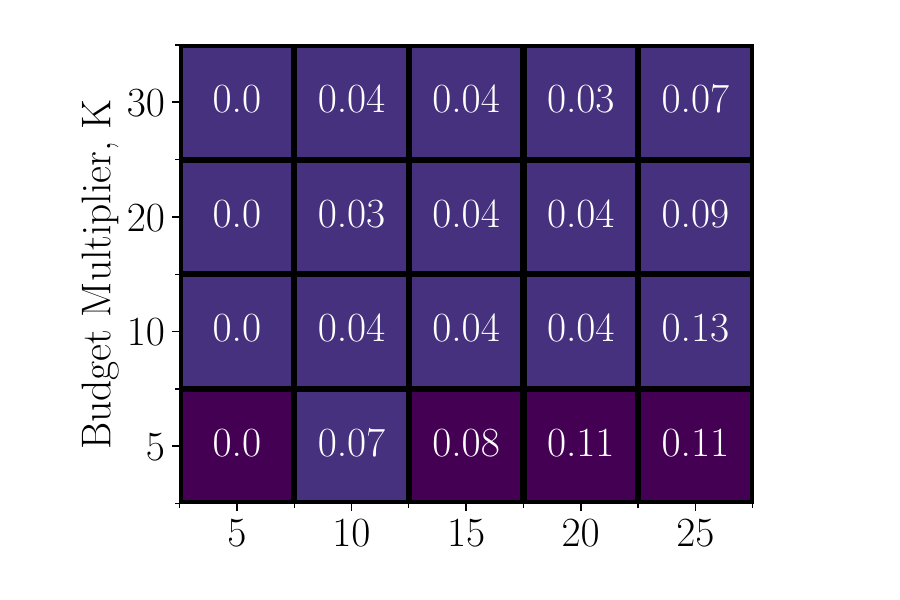} & \includegraphics[width=5cm]{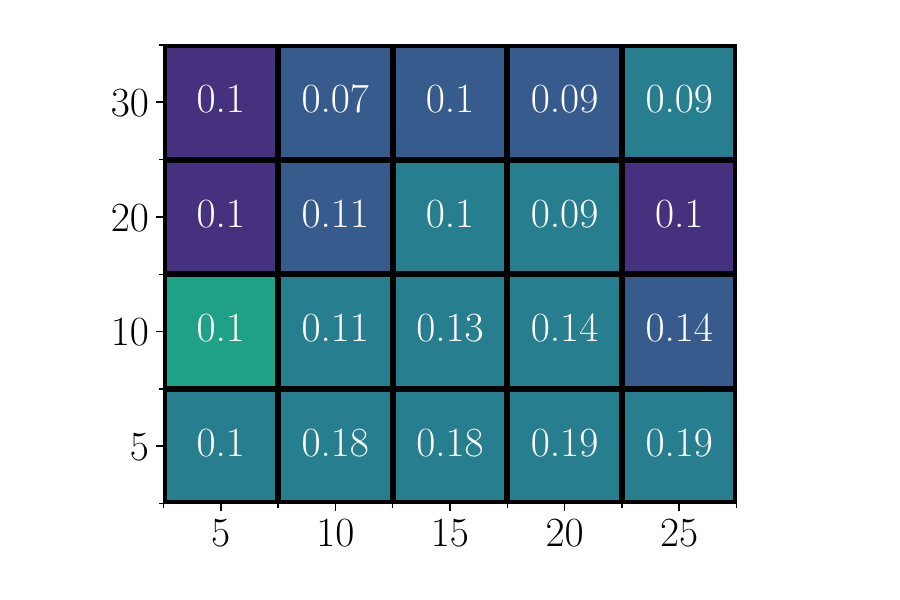} & \includegraphics[width=5cm]{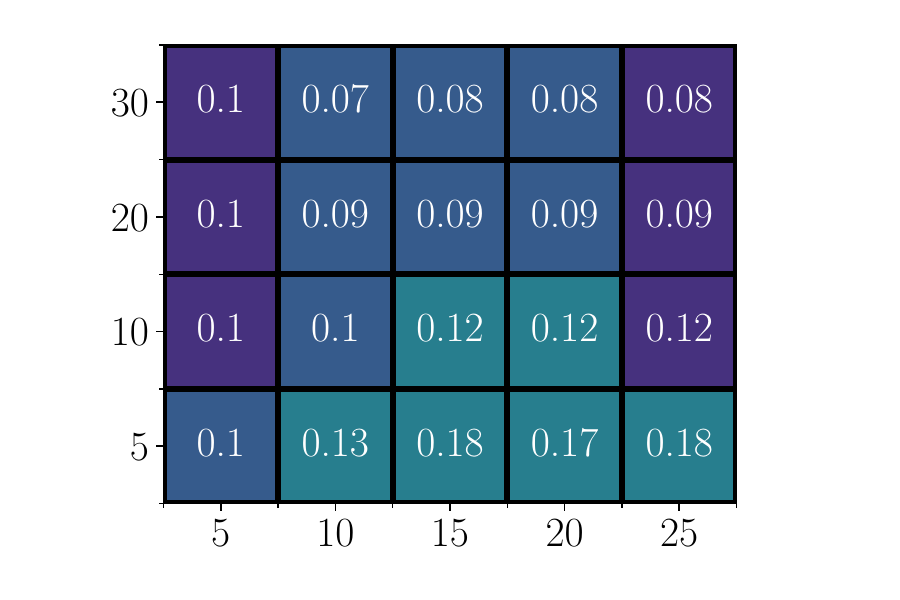} \\
                \hline
                \raisebox{50pt}[0pt][0pt]{\rotatebox[origin=c]{90}{MCR}} & \includegraphics[width=5.5cm]{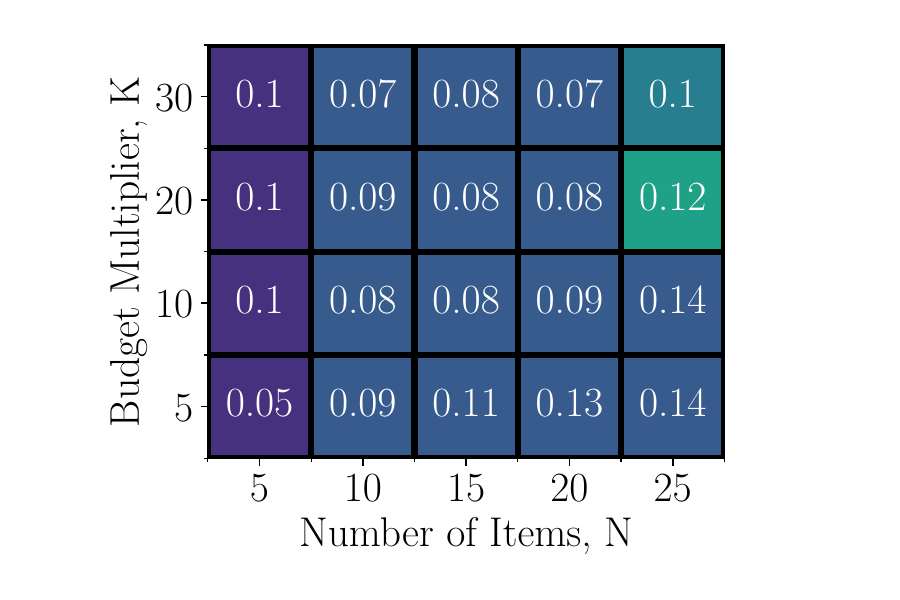} & \includegraphics[width=5.5cm]{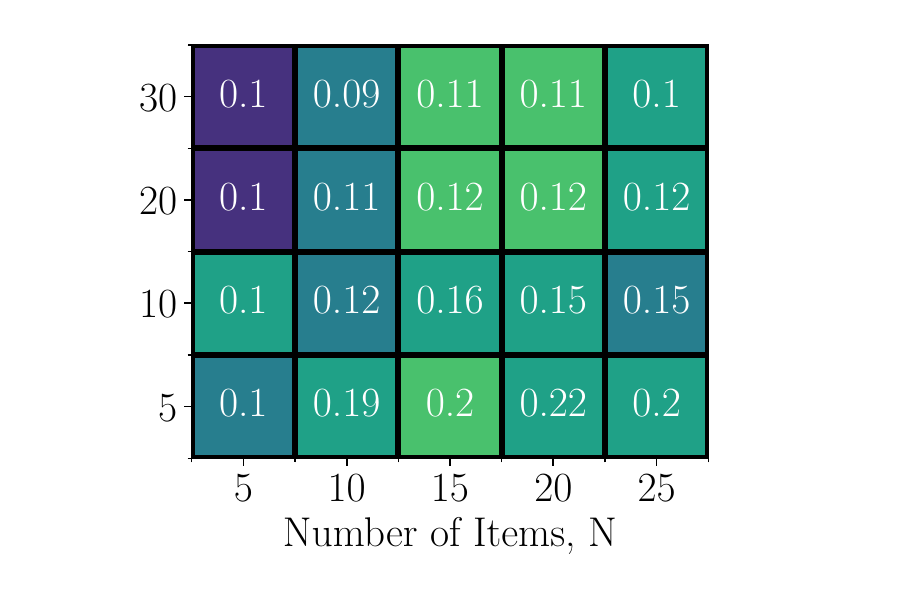} & \includegraphics[width=5.5cm]{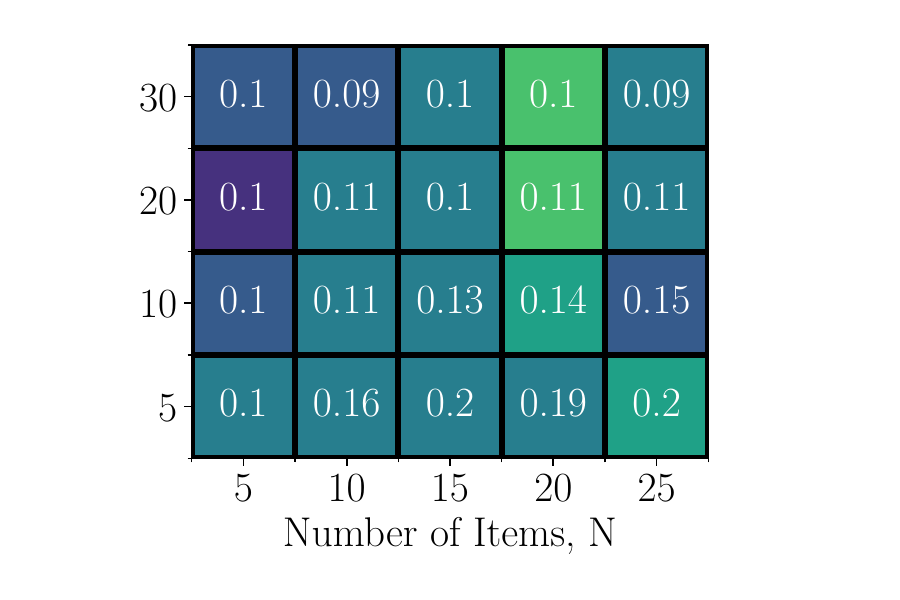} \\
                \hline
                \multicolumn{4}{c}{\includegraphics[width=6.5cm]{cplots/COLOURBAR_ONLY.pdf}} \\
            \end{tabular}   
            \caption{An illustration of the statistical comparison of multi-criteria strategies on the DREsS dataset using $50$ QMC-sampled weight vectors. The plots show that, overall, the MCP ranking method significantly outperforms the MCR ranking method. Among all strategies, the combination of MCP with entropy-based pair selection achieves the best performance. This winning strategy is only outperformed by the standard BCJ approach in one configuration.}
            \label{fig:wcrs_results_dress_random}
        \end{figure*}

\begin{figure}
            \centering
            \includegraphics[width=7cm]{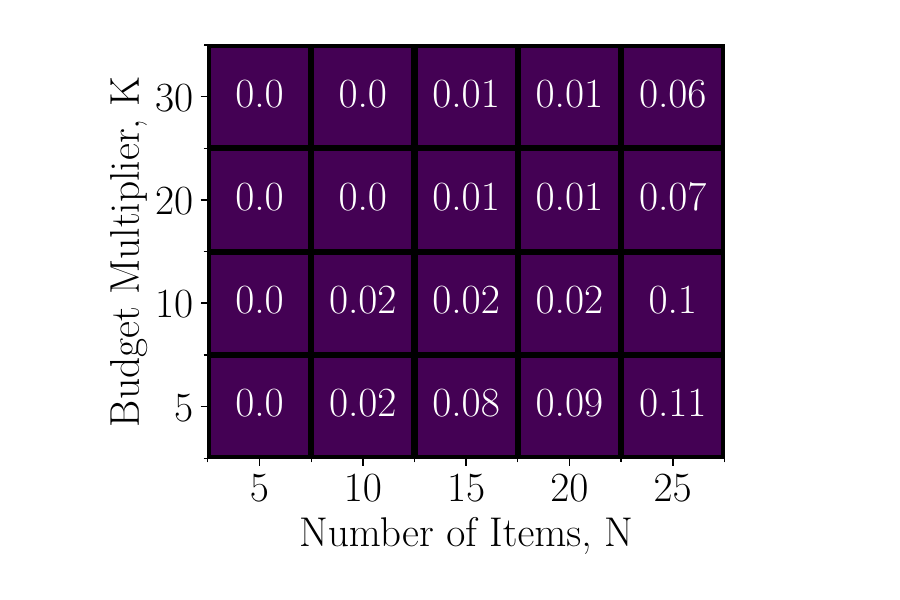}
            \caption{An illustration of the statistical comparison between the standard BCJ strategy and multi-criteria variants. Across all comparisons using $50$ QMC-sampled weight vectors, the BCJ strategy with entropy-based pair selection was not dominated by any other method. The strategy combining MCP with entropy also performed strongly, but was consistently outperformed by BCJ in these configurations.}
            \label{fig:bcj_dress_rw}
        \end{figure}

        In Figure~\ref{fig:wcrs_results_dress_random}, we present results from the DREsS dataset using 50 QMC-sampled weight vectors, independently generated for each $N$ and $K$ configuration, and strategy. Figure~\ref{fig:bcj_dress_rw} shows the corresponding outcomes for the standard single-criterion BCJ approach. Across all configurations, BCJ was not outperformed by any other strategy. However, the combination of MCP with entropy-based pair selection consistently achieved the best performance among the multi-criteria variants.

\begin{figure*}[]
            \centering
            \begin{tabular}{c c c c}
                \hline
                                    & Entropy    & Random   & No Repeating Pairs \\
                \hline
                \raisebox{50pt}[0pt][0pt]{\rotatebox[origin=c]{90}{MCP}} & \includegraphics[width=5cm]{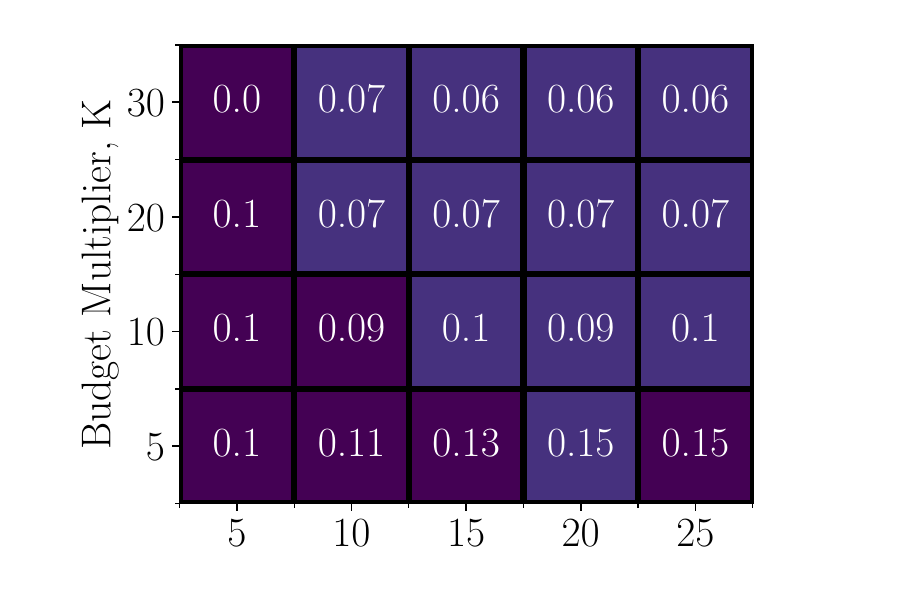}  & \includegraphics[width=5cm]{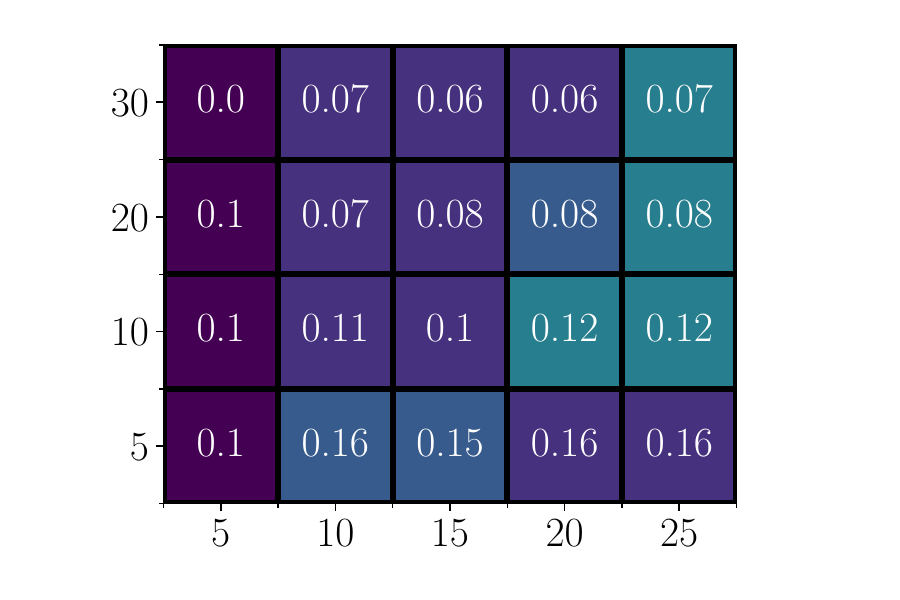}  & \includegraphics[width=5cm]{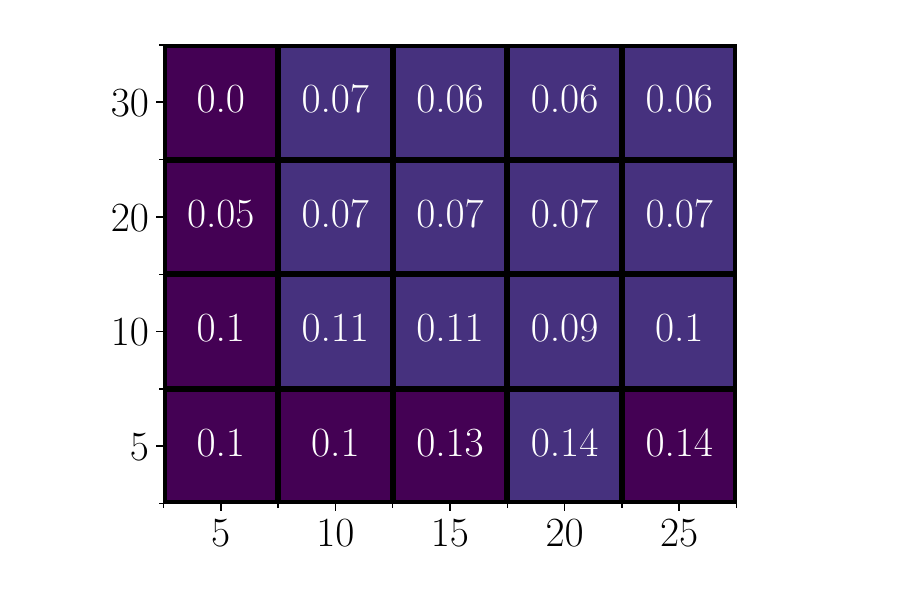} \\
                \hline
                \raisebox{50pt}[0pt][0pt]{\rotatebox[origin=c]{90}{MCR}} & \includegraphics[width=5.5cm]{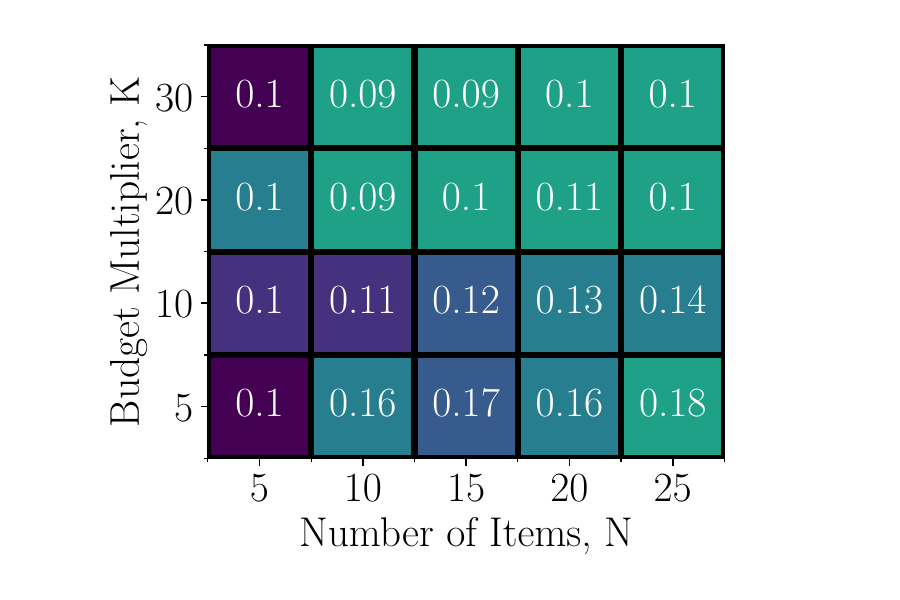} & \includegraphics[width=5.5cm]{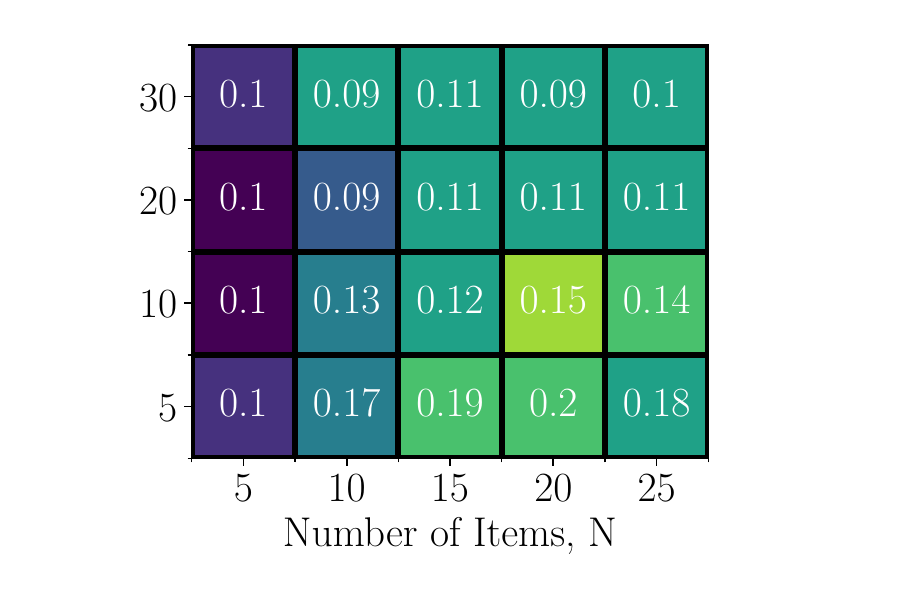} & \includegraphics[width=5.5cm]{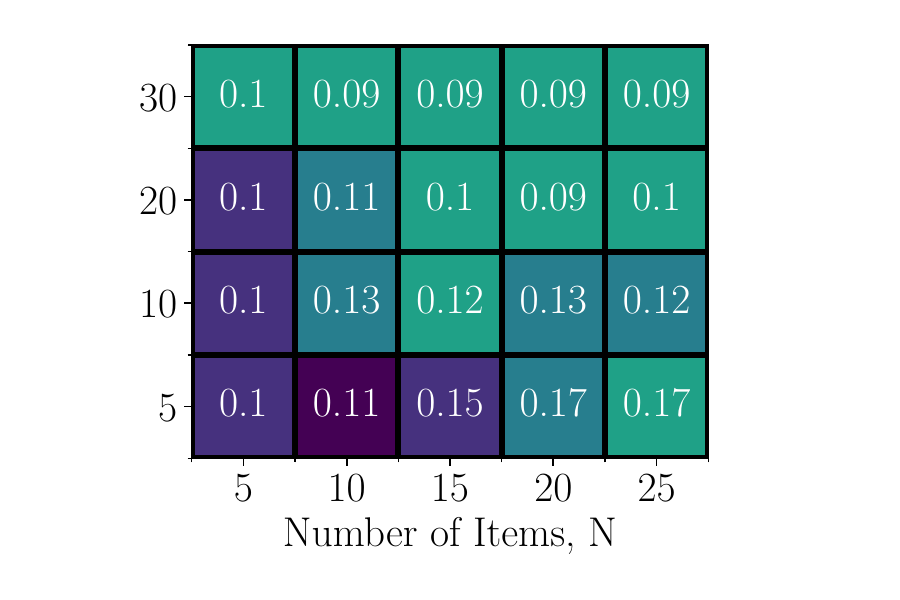} \\
                \hline
                \multicolumn{4}{c}{\includegraphics[width=6.5cm]{cplots/COLOURBAR_ONLY.pdf}} \\
            \end{tabular}   
            \caption{Statistical comparison of multi-criteria strategies for the BU dataset using QMC-sampled random weight vectors. The plots indicate that the MCP ranking method consistently outperformed the MCR approach. Among the MCP-based strategies, the pairing with the entropy selector showed a slight advantage over the combination with NRP for this dataset.}
            \label{fig:wcrs_results_random_bsu}
        \end{figure*}

\begin{figure}
            \centering
            \includegraphics[width=7cm]{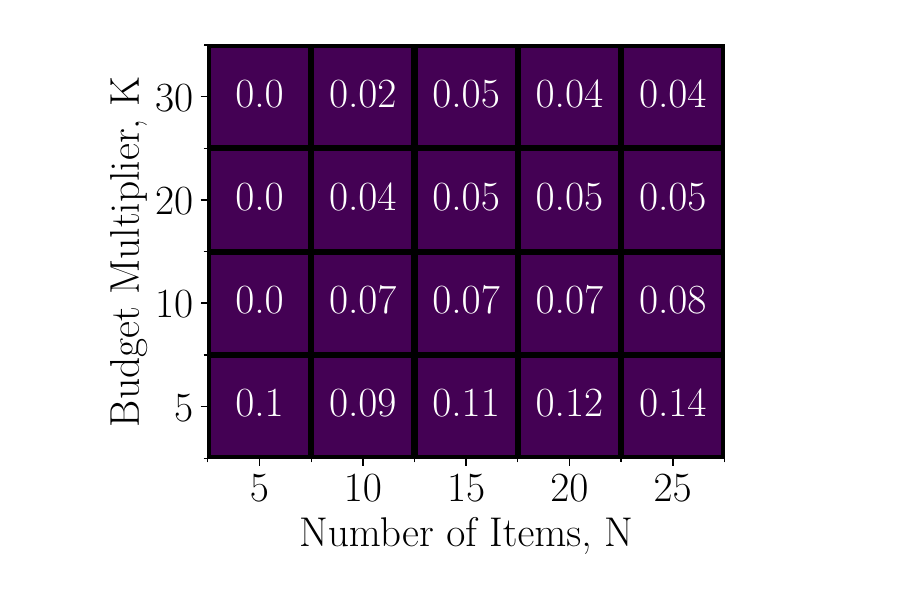}
            \caption{
            Statistical comparison of single-criterion BCJ combined with entropy-based pair selection versus multi-criteria variants, evaluated across 50 QMC-sampled weight vectors. The results show that BCJ with entropy selection was consistently competitive and not outperformed by any other method across the sampled weights.}
            \label{fig:wcrs_results_random_bcj_ent_bsu}
        \end{figure}

        The results for the BU dataset are presented in Figure \ref{fig:wcrs_results_random_bsu} (multi-criteria variants) and Figure \ref{fig:wcrs_results_random_bcj_ent_bsu} (single-criterion BCJ), and they closely mirror the findings from the DREsS dataset. Notably, the combination of MCP and entropy pair selectors performed slightly better here than in the DREsS results. A closer analysis reveals that this MCP–entropy pairing consistently outperformed other multi-criteria variants. MCP with NRP also showed strong performance, though not to the same extent. As observed across all experiments, MCR combined with any pair selector consistently underperformed compared to the MCP ranking method.

        In terms of robustness, BCJ emerges as the most effective method, with the MCP–entropy strategy following closely behind. This presents practitioners with a meaningful trade-off: BCJ offers a holistic and potentially faster decision-making process, requiring consideration of only a single comprehensive dimension. Conversely, multi-criteria approaches may take slightly longer, as each decision involves comparing all LOs within a pair. However, they offer richer insights into item discrimination, which can enhance feedback quality and improve transparency.

\subsection{Reassessing Scale Separation Reliability as a Metric}

    Across all experiments, our findings challenge the reliability of SSR as a definitive metric for ranking accuracy in CJ. We observed several cases where the target ranking was achieved (i.e. a $\tau$ score of 0), even though the corresponding SSR score fell well below the commonly recommended threshold of $0.7$. Conversely, higher SSR scores did not consistently lead to more accurate rankings. This suggests that SSR may be more reflective of the volume of comparisons — and the resulting confidence in those comparisons — rather than the actual quality of the final ranking.
    
    One particularly revealing insight was the wide variation in SSR scores across experiments. For instance, the lowest SSR score observed was $0.27$ for $N=5$, $K=5$, yet the $\tau$ score was $0.1$. The highest SSR score, $0.92$, occurred at $N=25$, $K=30$. Interestingly, in several cases where the $\tau$ score was 0 (indicating perfect ranking), the SSR score remained at $0.56$ — below the recommended threshold — for both $N=5$, $K=5$ and $N=5$, $K=30$. In another instance, with $N=5$, $K=30$, the SSR score was again $0.56$, but the $\tau$ score rose to $0.3$. Across $38$ experiments with an SSR score of $0.56$, $25$ achieved a $\tau$ score of 0.
    
    For example, when $N=5$ and $K=10$, the average SSR score was $0.56$, ranging from $0.47$ to $0.56$, yet 28 out of 50 runs resulted in a $\tau$ score of 0. Even at the lowest SSR score, the $\tau$ score was $0.1$, and at the highest SSR score of $0.56$, the $\tau$ score ranged from 0.0 to 0.1. These patterns suggest that SSR is more closely tied to the number of comparisons conducted, with higher SSR scores emerging from more extensive comparison sets. This raises important questions about the appropriateness of SSR as a measure of ranking accuracy in CJ, and highlights the need for further research into its role and limitations.
    
    While an SSR score provides a general sense of overall agreement, it does not reveal where assessors specifically disagreed. In contrast, MAP and EAP offer visual insights into which item pairs are contributing to disagreement, enabling a more targeted understanding of reliability, as demosntrated in Section \ref{sec:reliability-test}.

\section{Conclusion and Future Work}
    \label{sec:conclusion}

    Bayesian active learning for CJ introduces a new paradigm for efficiently collecting data through pairwise comparisons to produce accurate rankings. However, key limitations remain — notably, the lack of mechanisms to quantify decision reliability and the reliance on holistic comparisons rather than multi-criteria evaluations.
    
    This paper addresses these gaps with three core methodological contributions:
    \begin{itemize}
        \item \textbf{MAP and EAP:} Two novel reliability metrics that provide a fine-grained view of assessor agreement and enable targeted moderator interventions.
        \item \textbf{MCP:} A robust method for aggregating criterion-level judgements that maintains overall ranking accuracy while preserving detailed performance data across individual criteria.
        \item \textbf{Active Learning Strategy:} An efficient approach that reduces uncertainty across all criteria simultaneously.
    \end{itemize}
    
    In educational assessment, our multi-criteria BCJ framework enables both detailed feedback on individual LOs and an overall student ranking. When MCP ranking is combined with entropy-based pair selection, performance is comparable to standard holistic BCJ, while offering richer diagnostic insights.
    
    Experimental results show that MCP with entropy selection consistently performs well, often matching or surpassing the accuracy of holistic BCJ. Although holistic BCJ demonstrates greater robustness under marginalised criteria weights, our multi-criteria approach delivers fine-grained, actionable feedback with minimal impact on ranking accuracy. This makes it a valuable tool for formative assessment, allowing educators to understand student performance in depth without compromising reliability.

    \textbf{Future Work:} We plan to conduct user studies with educators and assessors to evaluate the impact of multi-criteria BCJ on ranking accuracy and cognitive load.
    
    It should be noted that both BCJ and multi-criteria BCJ are generalisable to a range of applications. For instance, the work of \cite{wainer2022bayesian} applied Comparative Judgement using a Bayesian Bradley–Terry model to evaluate classifiers based on a single accuracy criterion. A direct comparison with BCJ could be performed in this context.
    
    In terms of multi-criteria evaluation, there are currently no known examples that apply CJ to compare machine learning models across multiple metrics. However, multi-criteria decision analysis has been used for model comparison in this context — see, for example, \cite{akinsola2019performance}. This suggests a promising direction for future research: applying multi-criteria BCJ to evaluate classifiers using metrics such as true positive rate and false positive rate, possibly with equal weighting, to actively identify the best performing model in a data-efficient manner.

\section{Acknowledgements}
    \label{sec:ack}
    [name] 
    is funded by the EPSRC Centre for Doctoral Training in {\emph{Enhancing Human Interactions and Collaborations with Data and Intelligence-Driven Systems}} (EP/S021892/1) at 
    Swansea University. 
    Additionally, the project stakeholder is 
    CDSM. 
    We are particularly grateful to their CTO, 
    Darren Wallace. 
    We gratefully acknowledge 
    Saman Jayasinghe 
    for implementing the Bayesian Bradley-Terry model and providing valuable insights into its runtime performance. We would also like to thank 
    Jennifer Pearson 
    for their valuable feedback.
    
    For the purpose of Open Access, the author has applied a CC-BY public copyright licence to any Author Accepted Manuscript (AAM) version arising from this submission.
    All underlying data to support the conclusions are provided within this paper.

\appendix

\bibliographystyle{elsarticle-num} 
\balance
\bibliography{cas-refs}

\begin{thebibliography}{10}
\expandafter\ifx\csname url\endcsname\relax
  \def\url#1{\texttt{#1}}\fi
\expandafter\ifx\csname urlprefix\endcsname\relax\def\urlprefix{URL }\fi
\expandafter\ifx\csname href\endcsname\relax
  \def\href#1#2{#2} \def\path#1{#1}\fi

\bibitem{thurstone1927law}
L.~L. Thurstone, A law of comparative judgment, Psychological Review 34~(4) (1927) 273--286.
\newblock \href {https://doi.org/10.1037/h0070288} {\path{doi:10.1037/h0070288}}.

\bibitem{laming1984relativity}
D.~Laming, The relativity of ‘absolute’judgements, British Journal of Mathematical and Statistical Psychology 37~(2) (1984) 152--183.

\bibitem{bradley1952rank}
R.~A. Bradley, M.~E. Terry, {Rank analysis of incomplete block designs: The method of paired comparisons}, Biometrika 39~(3-4) (1952) 324--345.
\newblock \href {https://doi.org/10.1093/biomet/39.3-4.324} {\path{doi:10.1093/biomet/39.3-4.324}}.

\bibitem{hunter2004mm}
D.~R. Hunter, {MM algorithms for generalized Bradley-Terry models}, {Annals of Statistics} 32~(1) (2004) 384--406.
\newblock \href {https://doi.org/10.1214/aos/1079120141} {\path{doi:10.1214/aos/1079120141}}.

\bibitem{caron2012efficient}
F.~Caron, A.~Doucet, Efficient bayesian inference for generalized bradley--terry models, Journal of Computational and Graphical Statistics 21~(1) (2012) 174--196.

\bibitem{ballinger1997decisions}
T.~P. Ballinger, N.~T. Wilcox, Decisions, error and heterogeneity, The Economic Journal 107~(443) (1997) 1090--1105.

\bibitem{kelly2022critiquing}
K.~T. Kelly, M.~Richardson, T.~Isaacs, Critiquing the rationales for using comparative judgement: a call for clarity, Assessment in Education: Principles, Policy \& Practice 29~(6) (2022) 674--688.

\bibitem{pollitt2012method}
A.~Pollitt, The method of adaptive comparative judgement, {Assessment in Education: Principles, Policy \& Practice} 19~(3) (2012) 281--300.

\bibitem{bramley2015investigating}
T.~Bramley, {Investigating the reliability of Adaptive Comparative Judgment}, Tech. rep., {Cambridge Assessment} (March 2015).

\bibitem{jamieson2011active}
K.~G. Jamieson, R.~Nowak, Active ranking using pairwise comparisons, Advances in neural information processing systems 24 (2011).

\bibitem{heckel2019active}
R.~Heckel, N.~B. Shah, K.~Ramchandran, M.~J. Wainwright, Active ranking from pairwise comparisons and when parametric assumptions do not help (2019).

\bibitem{tversky1969substitutability}
A.~Tversky, J.~E. Russo, Substitutability and similarity in binary choices, Journal of Mathematical psychology 6~(1) (1969) 1--12.

\bibitem{GRAY2024100245}
A.~Gray, A.~Rahat, T.~Crick, S.~Lindsay, \href{https://www.sciencedirect.com/science/article/pii/S2666920X24000481}{A bayesian active learning approach to comparative judgement within education assessment}, Computers and Education: Artificial Intelligence (2024) 100245\href {https://doi.org/https://doi.org/10.1016/j.caeai.2024.100245} {\path{doi:https://doi.org/10.1016/j.caeai.2024.100245}}.
\newline\urlprefix\url{https://www.sciencedirect.com/science/article/pii/S2666920X24000481}

\bibitem{ashton2000review}
R.~H. Ashton, A review and analysis of research on the test--retest reliability of professional judgment, Journal of Behavioral Decision Making 13~(3) (2000) 277--294.

\bibitem{kinnear2025comparative}
G.~Kinnear, I.~Jones, B.~Davies, Comparative judgement as a research tool: a meta-analysis of application and reliability, Tech. rep., Center for Open Science (2025).

\bibitem{yu2022multidimensional}
Q.~Yu, K.~M. Quinn, A multidimensional pairwise comparison model for heterogeneous perceptions with an application to modelling the perceived truthfulness of public statements on covid-19, Journal of the Royal Statistical Society: Series A (Statistics in Society) 185~(3) (2022) 1049--1073.

\bibitem{hefner1959extensions}
R.~A. Hefner~Jr, Extensions of the law of comparative judgment to discriminable and multidimensional stimuli, University of Michigan, 1959.

\bibitem{wellington2007secondary}
J.~Wellington, Secondary education: The key concepts, Routledge, 2007.

\bibitem{yeomans2013teaching}
J.~Yeomans, C.~Arnold, Teaching, Learning and psychology, Routledge, 2013.

\bibitem{brooks2019preparing}
I.~Abbott, P.~Huddleston, D.~Middlewood, Preparing to teach in secondary schools: a student teacher's guide to professional issues in secondary education, McGraw-Hill Education (UK), 2012.

\bibitem{Cox2015The}
G.~Cox, J.~Morrison, B.~Brathwaite, The rubric: An assessment tool to guide students and markers, Headache (2015) 26--32\href {https://doi.org/10.4995/HEAD15.2015.414} {\path{doi:10.4995/HEAD15.2015.414}}.

\bibitem{Poh2015A}
B.~L.~G. Poh, K.~Muthoosamy, C.~Lai, G.~B. Hoe, A marking scheme rubric: To assess students' mathematical knowledge for applied algebra test, Asian Social Science 11 (2015) 18.
\newblock \href {https://doi.org/10.5539/ASS.V11N24P18} {\path{doi:10.5539/ASS.V11N24P18}}.

\bibitem{Ragupathi2020Beyond}
K.~Ragupathi, A.~Lee, Beyond fairness and consistency in grading: The role of rubrics in higher education, Diversity and inclusion in global higher education: Lessons from across Asia (2020) 73--95.

\bibitem{olson2021rubrics}
J.~M. Olson, R.~Krysiak, Rubrics as tools for effective assessment of student learning and program quality, in: Curriculum Development and Online Instruction for the 21st Century, IGI Global, 2021, pp. 173--200.

\bibitem{cox2015rubric}
G.~Cox, J.~Morrison, B.~Brathwaite, The rubric: an assessment tool to guide students and markers, in: 1ST INTERNATIONAL CONFERENCE ON HIGHER EDUCATION ADVANCES (HEAD'15), Editorial Universitat Polit{\`e}cnica de Val{\`e}ncia, 2015, pp. 26--32.

\bibitem{sambell2019assessment}
K.~Sambell, S.~Brown, P.~Race, Assessment to support student learning: eight challenges for 21st century practice, All Ireland Journal of Teaching and Learning in Higher Education (AISHE-J) Creative Commons Attribution-NonCommercial-ShareAlike 11~(2) (2019).

\bibitem{Jonsson2007The}
A.~Jonsson, G.~Svingby, The use of scoring rubrics: Reliability, validity, and educational consequences, Educational Research Review 2 (2007) 130--144.
\newblock \href {https://doi.org/10.1016/J.EDUREV.2007.05.002} {\path{doi:10.1016/J.EDUREV.2007.05.002}}.

\bibitem{Cockett2018The}
A.~Cockett, C.~Jackson, The use of assessment rubrics to enhance feedback in higher education: An integrative literature review., Nurse education today 69 (2018) 8--13.
\newblock \href {https://doi.org/10.1016/j.nedt.2018.06.022} {\path{doi:10.1016/j.nedt.2018.06.022}}.

\bibitem{Reddy2010A}
Y.~M. Reddy, H.~Andrade, A review of rubric use in higher education, Assessment \& Evaluation in Higher Education 35 (2010) 435 -- 448.
\newblock \href {https://doi.org/10.1080/02602930902862859} {\path{doi:10.1080/02602930902862859}}.

\bibitem{Panadero2020A}
E.~Panadero, A.~Jonsson, A critical review of the arguments against the use of rubrics, Educational Research Review 30 (2020) 100329.
\newblock \href {https://doi.org/10.1016/j.edurev.2020.100329} {\path{doi:10.1016/j.edurev.2020.100329}}.

\bibitem{Smit2017Effects}
R.~Smit, P.~Bachmann, V.~Blum, T.~Birri, K.~Hess, Effects of a rubric for mathematical reasoning on teaching and learning in primary school, Instructional Science 45 (2017) 603--622.
\newblock \href {https://doi.org/10.1007/S11251-017-9416-2} {\path{doi:10.1007/S11251-017-9416-2}}.

\bibitem{Hack2015Analytical}
C.~Hack, Analytical rubrics in higher education: A repository of empirical data, British Journal of Educational Technology 46~(5) (2015) 924--927.

\bibitem{chen-et-al:2023}
O.~Chen, F.~Paas, J.~Sweller, {A Cognitive Load Theory Approach to Defining and Measuring Task Complexity Through Element Interactivity}, {Educational Psychology Review} 35~(63) (2023).
\newblock \href {https://doi.org/10.1007/s10648-023-09782-w} {\path{doi:10.1007/s10648-023-09782-w}}.

\bibitem{sadler:1989}
D.~R. Sadler, Formative assessment and the design of instructional systems, {Instructional Science} 18 (1989) 119--144.
\newblock \href {https://doi.org/10.1007/BF00117714} {\path{doi:10.1007/BF00117714}}.

\bibitem{bramleypaired:2007}
T.~Bramley, Paired comparison methods, in: {Techniques for monitoring the comparability of examination standards}, 2007, pp. 246--300.

\bibitem{benton2018comparative}
T.~Benton, T.~Gallagher, Is comparative judgement just a quick form of multiple marking, {Research Matters: A Cambridge Assessment Publication} 26 (2018) 22--28.

\bibitem{bartholomew2019using}
S.~R. Bartholomew, G.~J. Strimel, E.~Yoshikawa, Using adaptive comparative judgment for student formative feedback and learning during a middle school design project, {International Journal of Technology and Design Education} 29~(2) (2019) 363--385.
\newblock \href {https://doi.org/10.1007/s10798-018-9442-7} {\path{doi:10.1007/s10798-018-9442-7}}.

\bibitem{christodoulou2017making}
D.~Christodoulou, {Making Good Progress?: The future of Assessment for Learning}, {Oxford University Press}, 2017.

\bibitem{pollitt1996raters}
A.~Pollitt, N.~L. Murray, What raters really pay attention to, {Studies in Language Testing} 3 (1996) 74--91.

\bibitem{pollitt2004let}
A.~Pollitt, Let’s stop marking exams, in: IAEA Conference, 2004, {University of Cambridge Local Examinations Syndicate}.

\bibitem{gray2022using}
A.~Gray, A.~A. Rahat, T.~Crick, S.~Lindsay, D.~Wallace, {Using Elo rating as a metric for comparative judgement in educational assessment}, in: Proceedings of 6th International Conference on Education and Multimedia Technology (ICEMT 2022), 2022, pp. 272--278.
\newblock \href {https://doi.org/10.1145/3551708.3556204} {\path{doi:10.1145/3551708.3556204}}.

\bibitem{luce1959individual}
R.~D. Luce, Individual choice behavior (1959).

\bibitem{andrich1978rating}
D.~Andrich, A rating formulation for ordered response categories, Psychometrika 43~(4) (1978) 561--573.
\newblock \href {https://doi.org/10.1007/BF02293814} {\path{doi:10.1007/BF02293814}}.

\bibitem{verhavert2018scale}
S.~Verhavert, S.~De~Maeyer, V.~Donche, L.~Coertjens, {Scale Separation Reliability: What Does It Mean in the Context of Comparative Judgment?}, {Applied Psychological Measurement} 42~(6) (2018) 428--445.
\newblock \href {https://doi.org/10.1177/0146621617748321} {\path{doi:10.1177/0146621617748321}}.

\bibitem{steedle2016evaluating}
J.~T. Steedle, S.~Ferrara, {Evaluating Comparative Judgment as an Approach to Essay Scoring}, {Applied Measurement in Education} 29~(3) (2016) 211--223.
\newblock \href {https://doi.org/10.1080/08957347.2016.1171769} {\path{doi:10.1080/08957347.2016.1171769}}.

\bibitem{hinkle2003applied}
D.~E. Hinkle, W.~Wiersma, S.~G. Jurs, {Applied Statistics for the Behavioural Sciences}, 6th Edition, {Houghton Mifflin}, 2002.

\bibitem{mcgrane2018applying}
J.~A. McGrane, S.~M. Humphry, S.~Heldsinger, Applying a thurstonian, two-stage method in the standardized assessment of writing, Applied Measurement in Education 31~(4) (2018) 297--311.

\bibitem{jones_davies_2022}
I.~Jones, B.~Davies, Comparative judgement in education research, {International Journal of Research \& Method in Education} (2022).
\newblock \href {https://doi.org/10.1080/1743727X.2023.2242273} {\path{doi:10.1080/1743727X.2023.2242273}}.

\bibitem{bergstra2012random}
J.~Bergstra, Y.~Bengio, Random search for hyper-parameter optimization, {Journal of Machine Learning Research} 13~(2) (2012) 281--305.

\bibitem{ofqual2017}
S.~Holmes, B.~Black, C.~Morin, Marking reliability studies 2017: Rank ordering versus marking – which is more reliable?, Tech. rep., Ofqual (January 2020).

\bibitem{wainer2022bayesian}
J.~Wainer, \href{http://jmlr.org/papers/v24/22-0907.html}{A bayesian bradley-terry model to compare multiple ml algorithms on multiple data sets}, Journal of Machine Learning Research 24~(341) (2023) 1--34.
\newline\urlprefix\url{http://jmlr.org/papers/v24/22-0907.html}

\bibitem{salvatier2016probabilistic}
J.~Salvatier, T.~V. Wiecki, C.~Fonnesbeck, Probabilistic programming in python using pymc3, PeerJ Computer Science 2 (2016) e55.

\bibitem{lewis1995sequential}
D.~D. Lewis, A sequential algorithm for training text classifiers: Corrigendum and additional data, {ACM SIGIR Forum} 29~(2) (1995) 13--19.
\newblock \href {https://doi.org/10.1145/219587.219592} {\path{doi:10.1145/219587.219592}}.

\bibitem{lazo1978entropy}
A.~V. Lazo, P.~Rathie, {On the entropy of continuous probability distributions (Corresp.)}, {IEEE Transactions on Information Theory} 24~(1) (1978) 120--122.
\newblock \href {https://doi.org/10.1109/TIT.1978.1055832} {\path{doi:10.1109/TIT.1978.1055832}}.

\bibitem{gray2023bayesian}
A.~Gray, A.~Rahat, T.~Crick, S.~Lindsay, A bayesian active learning approach to comparative judgement, arXiv preprint arXiv:2308.13292 (2023).

\bibitem{lindsay1995mixture}
B.~G. Lindsay, Mixture models: theory, geometry, and applications, Ims, 1995.

\bibitem{abramowitz1972stegun}
M.~Abramowitz, I. a. stegun (editors), handbook of mathematical functions, Applied Mathematics Series (1972).

\bibitem{korn2000mathematical}
G.~A. Korn, T.~M. Korn, Mathematical handbook for scientists and engineers: definitions, theorems, and formulas for reference and review, Courier Corporation, 2000.

\bibitem{yoo2024dress}
H.~Yoo, J.~Han, S.-Y. Ahn, A.~Oh, Dress: Dataset for rubric-based essay scoring on efl writing, arXiv preprint arXiv:2402.16733 (2024).

\bibitem{neyman1992two}
J.~Neyman, On the two different aspects of the representative method: the method of stratified sampling and the method of purposive selection, in: Breakthroughs in statistics: Methodology and distribution, Springer, 1992, pp. 123--150.

\bibitem{kendall1938new}
M.~G. Kendall, A new measure of rank correlation, Biometrika 30~(1-2) (1938) 81–93.
\newblock \href {https://doi.org/10.1093/biomet/30.1-2.81} {\path{doi:10.1093/biomet/30.1-2.81}}.

\bibitem{fagin2003comparing}
R.~Fagin, R.~Kumar, D.~Sivakumar, {Comparing Top k Lists}, {SIAM Journal on Discrete Mathematics} 17~(1) (2003) 134--160.
\newblock \href {https://doi.org/10.1137/S0895480102412856} {\path{doi:10.1137/S0895480102412856}}.

\bibitem{macfarland2016introduction}
T.~W. MacFarland, J.~M. Yates, et~al., Introduction to nonparametric statistics for the biological sciences using R, Springer, 2016.

\bibitem{dunn1961multiple}
O.~J. Dunn, Multiple comparisons among means, Journal of the American statistical association 56~(293) (1961) 52--64.

\bibitem{morokoff1995quasi}
W.~J. Morokoff, R.~E. Caflisch, Quasi-monte carlo integration, Journal of computational physics 122~(2) (1995) 218--230.

\bibitem{smith2004sampling}
N.~A. Smith, R.~W. Tromble, Sampling uniformly from the unit simplex, Johns Hopkins University, Tech. Rep 29 (2004).

\bibitem{akinsola2019performance}
J.~Akinsola, O.~Awodele, S.~Kuyoro, F.~Kasali, Performance evaluation of supervised machine learning algorithms using multi-criteria decision making techniques, in: Proceedings of the International Conference on Information Technology in Education and Development (ITED), 2019, pp. 17--34.

\end{thebibliography}

\bio{}
\endbio

\end{document}